\documentclass{article}

\usepackage[final]{automl}

\usepackage{natbib}

\usepackage[utf8]{inputenc} % allow utf-8 input
\usepackage[T1]{fontenc}    % use 8-bit T1 fonts
\usepackage{hyperref}       % hyperlinks
\usepackage{url}            % simple URL typesetting
\usepackage{booktabs}       % professional-quality tables
\usepackage{amsfonts}       % blackboard math symbols
\usepackage{nicefrac}       % ncompsact symbols for 1/2, etc.
\usepackage{microtype}      % microtypography
\usepackage{xcolor}         % colors
\usepackage{multirow}

\usepackage{mathtools}
\usepackage{graphicx}
\usepackage{amsmath}
\usepackage{booktabs}
\usepackage{algorithm}
\usepackage{algorithmic}

\usepackage[capitalise]{cleveref}
\usepackage{xspace}
\usepackage{centernot}
\usepackage{longtable}
\usepackage{caption}
\usepackage{enumitem}
\usepackage{xcolor}
\usepackage{wrapfig}
\usepackage{adjustbox}
\usepackage{algorithm}
\usepackage{algorithmic}
\usepackage{thmtools} 
\usepackage{thm-restate}
\usepackage[symbol]{footmisc}

\DeclareMathOperator*{\argmin}{arg\,min}

\makeatletter
\renewcommand\AB@affilsepx{, \protect\Affilfont}
\makeatother

\title{Automatic Termination for Hyperparameter Optimization} \looseness-1

\author[1]{\nameemail{Anastasiia Makarova}{anmakaro@ethz.ch}}
\author[2]{\nameemail{Huibin Shen}{huibishe@amazon.com}\footnote[1]{Correspondence to: Huibin Shen <huibishe@amazon.com>}}
\author[2]{\nameemail{Valerio Perrone}{vperrone@amazon.com}}
\author[2]{\nameemail{Aaron Klein}{kleiaaro@amazon.com}}
\author[2]{\nameemail{Jean Baptiste Faddoul}{faddoul@amazon.de}}
\author[1]{\nameemail{Andreas Krause}{krausea@ethz.ch}}
\author[2]{\nameemail{Matthias Seeger}{matthis@amazon.com}}
\author[2]{\nameemail{Cedric Archambeau}{cedrica@amazon.com}}

\affil[1]{ETH Z{ü}rich} 
\affil[2]{Amazon Web Services}

\begin{document}

\maketitle

%% Special characters for number sets.
\newcommand{\N}{\mathbb{N}}
\newcommand{\R}{\mathbb{R}}
\newcommand{\E}{\mathbb{E}}
\newcommand{\Normal}{\mathcal{N}}
\newcommand{\GP}{\mathcal{GP}}

%% Vectors and sets.
\newcommand{\set}[1]{\mathcal #1}
\newcommand{\mat}[1]{\mathbf #1}
\renewcommand{\vec}[1]{\mathbf #1}
\newcommand{\x}{\mathbf{x}}
\newcommand{\y}{\mathbf{y}}
\newcommand{\w}{\mathbf{w}}
\newcommand{\X}{\mathcal{X}}
\newcommand{\g}{\mathbf{\gamma}}
\newcommand{\D}{\mathcal{D}}
\newcommand{\ke}{\mathbf{k}}
\newcommand{\K}{\mathbf{K}}
\newcommand{\A}{\mathbf{A}}
\newcommand{\Y}{\mathbf{Y}}
\newcommand{\Si}{\mathbf{\Sigma}}
\newcommand{\eps}{\varepsilon}
\newcommand{\seps}{\sigma_{\varepsilon}}

%% Bold greek letters.
\newcommand{\lm}{\boldsymbol{\lambda}}
\newcommand{\tht}{\boldsymbol{\theta}}
\newcommand{\ph}{\boldsymbol{\phi}}
\newcommand{\omg}{\boldsymbol{\omega}}

%% Comments
\newcommand{\todo}[1] {\textcolor{red}{#1}}
\newcommand{\aaron}[1] {\textcolor{blue}{Aaron #1}}

%% Propositions
% \newtheorem{prop}{Proposition}

%% New
\newcommand{\Dtr}{\mathcal{D}_{tr}}
\newcommand{\Dtv}{\mathcal{D}_{tv}}
\newcommand{\Dval}{\mathcal{D}_{val}}
\newcommand{\Dtest}{\mathcal{D}_{G}}
\newcommand{\VarG}{\mathrm{Var}\festim}
\newcommand{\VarGt}{s^2_t}
\newcommand{\VarGset}{S^2}
\newcommand{\VarGincumbent}{\hat{s}^2}
\newcommand{\stdG}{s}
\newcommand{\VarCV}{s^2_{\mathrm{cv}}}
\newcommand{\VarCVt}{{s^2_{cv}}_t}
\newcommand{\Dt}{G_t}
\newcommand{\Dgamma}{\mathfrak D}
\newcommand{\simpleregret}{r}
\newcommand{\simpleregretestim}{\hat{r}}
\newcommand{\algo}{\mathcal{M}}
\newcommand{\festim}{\hat{f}}
\newcommand{\goptim}{\gamma^*}
\newcommand{\goptimD}{\gamma_{\D}^*}
\newcommand{\goptimDestim}{\gamma_{\D}}
\newcommand{\lcb}{\mathrm{lcb}\xspace}
\newcommand{\ucb}{\mathrm{ucb}\xspace}

\newcommand{\krepeat}{k}
\newcommand{\samplevar}{\hat{s}^2_{\krepeat}(x)}
\newcommand{\varproxy}{var}
\newcommand{\raucb}{\textsc{RAHBO}\xspace}
\newcommand{\gpucb}{\textsc{GP-UCB}\xspace}
\newcommand{\raucbus}{\textsc{RAHBO-US}\xspace}
\newcommand{\swissfel}{SwissFEL\xspace}
\newcommand{\xreported}{\hat{x}_T}
\newcommand{\mv}{\text{MV}\xspace}
\newcommand{\ubar}[1]{\text{\b{$#1$}}}
\newcommand{\rhomin}{\smash{\underline{\varrho}}}
\newcommand{\rhomax}{\bar{\varrho}}
\newcommand{\sigman}{$\sigma^2_{\varepsilon}$}

\begin{abstract}
%Tuning machine learning models with Bayesian optimization (BO) is a successful strategy to find good hyperparameters. BO defines an iterative procedure where a cross-validated metric is evaluated on promising hyperparameters. In practice, however, an improvement of the validation metric may not translate in better predictive performance on a test set, especially when tuning models trained on small datasets. In other words, unlike conventional wisdom dictates, BO can overfit. In this paper, we carry out the first systematic investigation of overfitting in BO and demonstrate that this issue is serious, yet often overlooked in practice. We propose a novel criterion to early stop BO, which aims to maintain the solution quality while saving the unnecessary iterations that  can lead to overfitting. Experiments on real-world hyperparameter optimization problems show that our approach effectively meets these goals and is more adaptive comparing to baselines. 

Bayesian optimization (BO) is a widely popular approach for the hyperparameter optimization (HPO) 
in machine learning. 
At its core, BO iteratively evaluates promising configurations until a user-defined budget, such as wall-clock time or number of iterations, is exhausted. While the final performance after tuning heavily depends on the provided budget, it is hard to pre-specify an optimal value in advance. In this work, we propose an effective and intuitive termination criterion for BO that automatically stops 
the procedure if it is 
sufficiently close to the global optimum. Our key insight is that the discrepancy between the true objective (predictive performance on test data) and the computable target (validation performance) suggests stopping once the suboptimality in optimizing the target is dominated by the statistical estimation error.
Across an extensive range of real-world HPO problems and baselines, we show that our termination criterion achieves a better trade-off between the test performance and optimization time.
Additionally, we find that overfitting may occur in the context of HPO, which is arguably an overlooked problem in the literature, and show how our termination criterion helps to mitigate this phenomenon on both small and large datasets. \looseness-1

%automatic termination baselines.  and that can be used in conjunction with any given HPO optimizer.
% Experiments on an extensive range of real-world hyperparameter optimization problems show that our termination criterion is more robust and maintains a higher solution quality compared to baselines.
\end{abstract}

\maketitle
\section{Introduction}
\label{sec:intro}

While the performance of machine learning algorithms crucially depends on their hyperparameters, setting them correctly is typically a tedious and expensive task. 
Hyperparameter optimization (HPO) emerged as a new sub-field in machine learning that tries to automatically determine how to configure a machine learning model.
One of the most successful strategies for HPO is Bayesian optimization  \citep[BO;][]{movckus1975,chen2018bayesian,snoek12,melis2018on} - a powerful framework for sequentially optimizing a costly blackbox objective, that is the predictive performance of a model configured with certain hyperparameters. BO iteratively searches for a better predictive performance via (i) training a probabilistic model on the evaluations
of the models performance and (ii) selecting the most promising next hyperparameter candidate.
% by trading off exploration and exploitation.

In practice, the quality of the solution found by BO heavily depends on a pre-defined budget, such as the number of BO iterations or wall-clock time. If this budget is too small, BO might result into hyperparameters of poor predictive performance. 
If the budget is too large, compute resources will be wasted. The latter can be especially fragile in HPO when one cannot fully reduce the discrepancy between the validation and test errors, thus resulting in {\em overfitting} as we show in our experiments. \looseness-1
%Automatically terminating the sequential procedure of BO is a rather under-explored topic, in contrast to the more widely considered orthogonal direction of speeding up HPO via stopping the model training   \citep{Klein2017LearningCP,training_earlystopping2019,li2017hyperband}. 

A naive approach suggests terminating BO if the best-found solution remains unchanged for some subsequent BO iterations. Though the idea is sensible, it might be challenging to define a suitable number, since it is a fixed, predetermined choice, that does not take the observed data into account. Another approach is to track the probability of improvement \citep{lorenz2016stopping} or the expected improvement \citep{Nguyen2017regret}, and stop the optimization process once it falls below a given threshold. However, determining this threshold may in practice be less intuitive than setting the number of iterations or the wall-clock time. Instead of stopping BO completely, in \cite{mcleod18a}, it is proposed to switch to local optimization when the global regret is smaller than a pre-defined target. This condition can also be used to terminate BO early, but it comes with additional complexity such as identifying a (convex) region for local optimization and again a predefined budget. \looseness-1

Automatically terminating the sequential procedure of BO is a rather under-explored topic, in contrast to the more widely considered orthogonal direction of speeding up HPO via stopping the model training. A seminal idea there is to avoid the computation of low performing hyperparameters, e.g., by learning curves \citep{Swersky2014FreezeThawBO}, multi-fidelity approach Hyperband \citep{li2017hyperband}, its combination \citep{Klein2017LearningCP}, and further modifications like Hyperband BOHB \citep{falkner2018bohb}, asynchronous Hyperband \citep{li2020system} and its model based version \citep{klein2020model}. The profound distinction of our method with this line of works is in the problem setup: instead of stopping the training, our method aims to terminate the whole HPO process. This orthogonality allows for combination of both ways of to achieve overall larger speed ups. \looseness-1

In this work, we propose a simple and interpretable automatic termination criterion for BO. 
The criterion consists of two main ingredients: (i) high-probability confidence bound on the regret (i.e., the difference of our current solution to the global optimum) and (ii) the termination threshold. The first already allows a user to specify a desired tolerance that defines how accurate should the final solution be compared to the global optimum. For the case when cross-validation is used, we recommend a threshold based on the statistical properties of the cross-validation estimator. This threshold takes into account the irreducible discrepancy between the actual HPO objective (i.e., performance on new data) and the target function optimized via BO (i.e., the validation error). Our extensive
empirical evaluation on a variety of HPO and neural architecture search (NAS) benchmarks suggests that our method is more robust and effective in maintaining the final solution quality than common baselines. We also surface overfitting effects in HPO on both small and large datasets, arguably
an overlooked problem, and demonstrate that our termination criterion helps to mitigate it.  \looseness-1

\section{Background}
\label{sec:BO} 

	\textbf{Bayesian optimization} (BO) refers to a class of methods
	for gradient-free optimization of an objective $f: \Gamma \rightarrow \R$ in an iterative manner. At every step $t$, a learner selects an input $\gamma_t \in \Gamma$ and observes a noisy output \looseness-1 
	\begin{align*}
	    y_t \triangleq f(\gamma_t) + \varepsilon_t,
	\end{align*}
	
% 	\begin{align*}
% 	    y_t = f(\gamma_t) + \varepsilon_t,
% 	\end{align*}
	
	where  $\varepsilon_t$ is typically assumed to be i.i.d.~(sub)-Gaussian noise with some variance  $\sigma_{\varepsilon}^2$. 
	%CA: why sub-Gaussian and why "proxy"? AM: because for sub-Gaussian noise (which is typical assumption that we mention)variance-proxy is used instead of variance
	The decision of the next input %point 
	to evaluate depends on a probabilistic  model, used to approximate the objective $f$, and an acquisition function, which determines the decision rule. A popular choice for the probabilistic model is a {\em Gaussian process (GP)}: $f \sim GP(\mu, \kappa)$ \citep{rasmussen2006}, specified by some mean function $\mu: \Gamma \rightarrow \R $ and some kernel $\kappa: \Gamma \times \Gamma \rightarrow \R$. As observations $y_{1:t} = [y_1, \dots, y_t]^\top$ for the selected inputs $\Dt = \{\gamma_1,\dots,\gamma_t\}$ are being collected, they are used to update the posterior belief of the model defined by the posterior mean $\mu_t(\gamma)$ and variance $\sigma_t^2(\gamma)$:% as:
	\begin{align}
	    &\mu_t(\gamma) = \kappa_t(\gamma)^T(K_t + \sigma^2_{\varepsilon} I)^{-1} y_{1:t}, &\sigma^2_t(\gamma) = \kappa(\gamma,\gamma) - \kappa_t(\gamma)^\top(K_t + \sigma^2_{\varepsilon}I)^{-1}\kappa_t(\gamma) \label{eq:posterior},
	\end{align}
	where $(K_t)_{i,j} = \kappa(\gamma_i, \gamma_j)$ and $\kappa_t(\gamma)^T = [\kappa(\gamma_1, \gamma), \dots, \kappa(\gamma_t, \gamma) ]^T$. The next input %point 
	to query is determined by an {\em acquisition function} that aims to trade off exploration and exploitation. Common choices include probability of improvement~\citep{kushner1963new}, entropy search~\citep{Hennig2012} and GP upper-confidence bound~\citep{srinivas10} to name a few.   The convergence of BO can be quantified via {\em (simple) regret}:%, i.e., the sub-optimality in function value: 
\begin{align}
&\simpleregret_t := f(\goptim_t) - f(\goptim), \label{eq:simpleregret}
\end{align}
where $\goptim$ is the global optimizer of $f$ and $\goptim_t = \argmin_{\gamma \in \Dt} f(\gamma)$. 
% Some BO algorithms are equipped with theoretical convergence guarantees for $\simpleregret_t$, providing (under certain regularity assumptions on $f$) bounds for the number of iterations $t$ required to achieve an $\epsilon$-near-optimal solution, i.e., $t: \ r_t \leq \epsilon$ for some given value $\epsilon \geq 0$ (e.g., \cite{bogunovic2016truncated}).
Specifying adequate tolerance that defines how small the regret should be to terminate BO is of high importance as it determines both the quality and the cost of the solution. However, this criterion cannot be directly evaluated in practice, as the input $\goptim$ and the optimum $f(\goptim)$ are not known. \looseness-1

\textbf{Hyperparameter optimization} (HPO) is a %widely considered
classical application for BO.  Consider a supervised learning %setting
problem %consisting in training 
that requires to train a machine learning model (e.g., a neural network) $\algo$ on some feature-response data points %CA: why not just "featured data points" 
$\D = \{ (\mathrm{x}_i,\mathrm{y}_i)\}_{i=1}^n$ sampled i.i.d.~from some unknown data distribution $P$. The model is obtained by running a training algorithm (e.g., optimizing the weights of the neural network via SGD) on $\D$, both of which depend on {\em hyperparameters} $\gamma$ (e.g., learning rates used, batch size, etc.).
We use the notation 
$\algo_{\gamma}(\mathrm{x} ; \D)$ to refer to the prediction that the model produced by $\mathcal{M}$ makes for an input $\mathrm{x}$, when trained with hyperparameters $\gamma$ on data $\D$.
Given some loss function $\ell(\cdot, \cdot)$, the {\em population risk} of the model on unseen data points is given by the expected loss $\E_{P} [\ell(\mathrm{y}, \algo_{\gamma}(\mathrm{x}, \D))]$. The main objective of HPO is to identify hyperparameters $\gamma$, such that the resulting model minimizes the population risk:
\begin{align}
   &f(\gamma) =  \E_{(\mathrm{x}, \mathrm{y}) \sim P} \big[\ell \big(\mathrm{y}, \algo_{\gamma}(\mathrm{x}, \D) \big) \big], & \gamma^* = \underset{\gamma \in \Gamma}{\argmin} f(\gamma). \label{eq:gen_bo_problem}
\end{align}
In practice, however, the population risk cannot be evaluated since $P$ is unknown.  Thus, typically, it is estimated on a separate finite validation set $\D_{V}$ drawn from the same distribution $P$.
%As a result, p
Practical HPO focuses on minimizing the {\em empirical estimator} $\festim(\gamma)$ of the expected loss $f(\gamma)$ leading to %(probably different) 
the optimizer $\goptimD$: \looseness-1
	\begin{align}
	&\festim(\gamma) = \frac{1}{|\D_V|} \underset{(\mathrm{x}_i, \mathrm y_i) \in \D_V}{\sum} \ell\big(\mathrm{y}_i, \algo_{\gamma}(\mathrm{x}_i, \D)\big), &\goptimD = \underset{\gamma \in \Gamma}{\argmin} \festim(\gamma). 
 	\label{eq:val_bo_problem}
	\end{align}

At its core, BO-based HPO sequentially evaluates the empirical estimator $\festim(\gamma_t)$ for promising hyperparameters $\gamma_t$ and terminates after some specified number of BO rounds, reporting the solution 
$\goptim_t = \underset{\gamma \in \Dt}{\argmin} \festim(\gamma)$, where $\Dt=\{\gamma_1,\dots,\gamma_t\}$ are the solutions considered so far. We can define the simple regret for the reported solution w.r.t.~the validation loss by   \looseness-1
\begin{align}
&\simpleregretestim_t := \festim(\goptim_{t}) - \festim(\goptimD). \label{eq:simpleregretestim}%CA: I assume you mean $\goptimD$ here and not $\goptim$?
\end{align}

% Analogously, the sequence of $\{\simpleregretestim_1, \dots \simpleregretestim_t\}$ quantifies the convergence of BO measuring sub-optimality of the observed solutions. Intuitively, larger $t$ should lead to smaller  $r_t$, however these values are again not available in practice.
% Nevertheless,  and one might be interested to insure that it is smaller than some $\epsilon_{BO}$, but e. 

\textbf{Inconsistency in the optimization objective.} Importantly, the true HPO objective $f(\gamma)$ in \cref{eq:gen_bo_problem}
and the empirical surrogate $\festim(\gamma)$ in \cref{eq:val_bo_problem} used for tuning by BO generally do not coincide. Therefore, existing BO approaches may yield sub-optimal solutions to the population risk minimization, even if they succeed in globally optimizing $\festim(\gamma)$. This issue, however, is typically neglected in practical HPO as well as a potential overfitting to the validation error.  In contrast, we propose a termination condition for BO motivated by the discrepancy in the objectives.  \looseness-1

%\newpage
\section{Termination criterion for Hyperparameter Optimization}
\label{sec:main}

This section firstly motivates why early termination of HPO can be beneficial and then addresses the following two questions: (1) How to estimate the unknown simple regret and (2) What threshold of the simple regret can be used to stop HPO.

\subsection{Motivation for the termination criterion} \label{sec:motivation}

We start by analysing the effect of optimizing $\festim$ in lieu of $f$.
%true discrepancy of interest: the discrepancy between the population risk $f(\goptim_{t})$ %at some input  $\goptim_{t}$ 
%and the true optimum $f(\goptim)$. We %then 
We observe that challenges in optimizing $f$ are both due to
%this discrepancy sums from 
the statistical error of the empirical BO objective $\festim(\gamma)$ %as well 
and the sub-optimality of the BO candidates, encoded in the simple regret $\simpleregretestim_t$. The key insight of the following proposition is that iteratively reducing $\simpleregretestim_t$ to 0 may not bring any benefits if the statistical error dominates. \looseness-1

% by observing that the regret in terms of expected loss $f$ depends both on the statistical error $\epsilon_{st}$ of an empirical estimate  $\festim$ as well as the optimization error $\epsilon_{BO}$  $\festim$. The key insight is that reducing $\epsilon_{BO}$ to 0 may not bring any benefits if the statistical error $\epsilon_{st}$ dominates. \looseness-1  %%CA: do you mean error?  % DONE

\textbf{Proposition 1.}  \textit{Consider the expected loss $f$  and its estimator $\festim$ defined, respectively, in \cref{eq:gen_bo_problem,eq:val_bo_problem}, and assume the statistical error of the estimator is bounded as $\|\festim-f\|_\infty \leq\epsilon_{st}$ for some $\epsilon_{st} \geq 0$.
Let $\goptim$ and $\goptimD$ be their optimizers: $\goptim = \argmin_{\gamma \in \Gamma} f(\gamma)$ and $\goptimD = \argmin_{{\gamma \in \Gamma}} \festim(\gamma)$. Let $\goptim_{t}$ be some candidate solution to $\min_{\gamma \in \Gamma} \festim(\gamma)$ 
with sub-optimality in function value $\simpleregretestim_t := \festim(\goptim_{t}) - \festim(\goptimD)$. Then the gap in generalization performance $ f(\goptim_{t}) - f(\goptim)$ 
% between the proposed solution and the global optimum 
can be bounded as follows:}
\begin{align}
    f(\goptim_{t}) - f(\goptim) 
    &=\underbrace{f(\goptim_{t}) - \festim(\goptim_{t})}_{\leq \epsilon_{st}}
    + \underbrace{\festim(\goptim_{t}) - \festim (\goptimD)}_{= \simpleregretestim_t} 
    + \underbrace{\festim (\goptimD) - \festim(\goptim)}_{\leq 0}
    + \underbrace{\festim (\goptim) - f(\goptim)}_{\leq \epsilon_{st}} \label{eq:prop1_eq}\\
    &\leq 2 \epsilon_{st}+\simpleregretestim_t. \label{eq:prop1_ineq}
\end{align}
\vspace{-0.1pt}
{\em Moreover, without further restrictions on $f$, $\festim$, $\goptim_{t}$ and $\goptim$, the upper bound is tight.} \looseness-1

\textit{Proof:}
The equality in \cref{eq:prop1_eq} is due to adding and subtracting the same values.  The inequality in \cref{eq:prop1_ineq} results from the following bounds: \looseness-1
\begin{align}
    & (1) \  f(\goptim_{t}) - \festim(\goptim_{t}) \leq |f(\goptim_{t}) - \festim(\goptim_{t}) |\leq \max_{\gamma \in \Gamma} |f(\gamma) - \festim(\gamma)| = ||\festim-f||_\infty \leq\epsilon_{st}, \nonumber \\
    & (2)\  \goptimD = \argmin_{{\gamma \in \Gamma}} \festim(\gamma) \ \ \ \ \Rightarrow \ \ \ \ \forall \gamma \in \Gamma: \festim (\goptimD) - \festim(\gamma)   \leq 0 \ \ \ \ \Rightarrow \ \ \ \ \festim (\goptimD) - \festim(\goptim) \leq 0. \nonumber \ \ \  \blacksquare
\end{align}
    The proposition bounds the sub-optimality of the target objective $f$ in terms of the statistical error $\epsilon_{st}$ and the simple regret $\simpleregretestim_t$. This naturally suggests terminating HPO at a candidate $\goptim_{t}$ for which the simple regret $\simpleregretestim_t$ is of the same magnitude as the statistical error $\epsilon_{st}$, \looseness-1 
    since further reduction in $\simpleregretestim_t$ may not improve notably the true objective. However, neither of the quantities $\epsilon_{st}$ and $\simpleregretestim_t$ are known. \looseness-1
% A natural idea is thus to terminate HPO at solution $\gamma_t$ once the optimization error $\epsilon_{BO}$, which is bounded by the simple regret $\simpleregretestim_t$, is of the same magnitude as the statistical error $\epsilon_{st}$. 
% Unfortunately, neither of these quantities is known.  

Below, we propose a termination criterion that relies on estimates of both quantities.  Firstly, we show how to use confidence bounds on $\festim(\gamma)$ to obtain high probability upper bounds on the simple regret $\simpleregretestim_t$ \citep{srinivas10,ha_unknown_search_bo}. 
Secondly, we estimate the statistical error $\epsilon_{st}$ in the case of cross-validation \citep{stone1974cross,geisser75} where the model performance is defined as an average over several training-validation runs. To this end, we rely on the statistical characteristics (i.e., variance or bias) of such cross-validation-based estimator that are theoretically studied by \cite{nadeau2003inference} and \cite{bayle2020crossvalidation}. 
% When cross-validation is not used, we can define an intuitive threshold in advance based on the usage of an interpretable upper bound of the simple regret. \looseness-1

\subsection{Building blocks of the termination criterion} \label{sec:termination}

\textbf{Upper bound for the simple regret $\simpleregretestim_t$. }
The key idea behind bounding $\simpleregretestim_t$ is that, as long as the GP-based %surrogate model 
approximation of $\festim(\cdot)$ is well-calibrated, we can use it to construct high-probability confidence bounds for $\festim(\cdot)$.
In particular, \citet{srinivas10} show that, as long as $\festim$ has a bounded norm in the reproducing kernel Hilbert space (RKHS) associated with the covariance function $\kappa$ used in the GP, $\festim(\gamma)$ is bounded (with high probability) by lower and upper confidence bounds  $\lcb_t(\gamma) = \mu_t(\gamma) - \sqrt{\beta_t}\sigma_t(\gamma)$ and $\ucb_t(\gamma) = \mu_t(\gamma) + \sqrt{\beta_t}\sigma_t(\gamma)$. Hereby, $\beta_t$ is a parameter that ensures validity of the confidence bounds (see \cref{app:beta_choice} for practical discussion and ablation study).\looseness -1 

% Clearly, $\beta_t$ impacts the quality of the upper bound significantly and throughout this paper, we set $\beta_t = 2 \log (|\Gamma| t^2\pi^2/6\delta)$ where $\delta=0.1$ and $|\Gamma|$ is set to be the number of hyperparameters. We then further scale it down by a factor of 5 as defined in the experiments in \cite{srinivas10}. We provide an ablation study on the choice of $\beta_t$ in \cref{fig:betat} in the appendix. 
%chosen to ensure the validity of confidence bounds.\looseness -1

\begin{figure}[t!]
 \centering
 \includegraphics[width=0.45\textwidth]{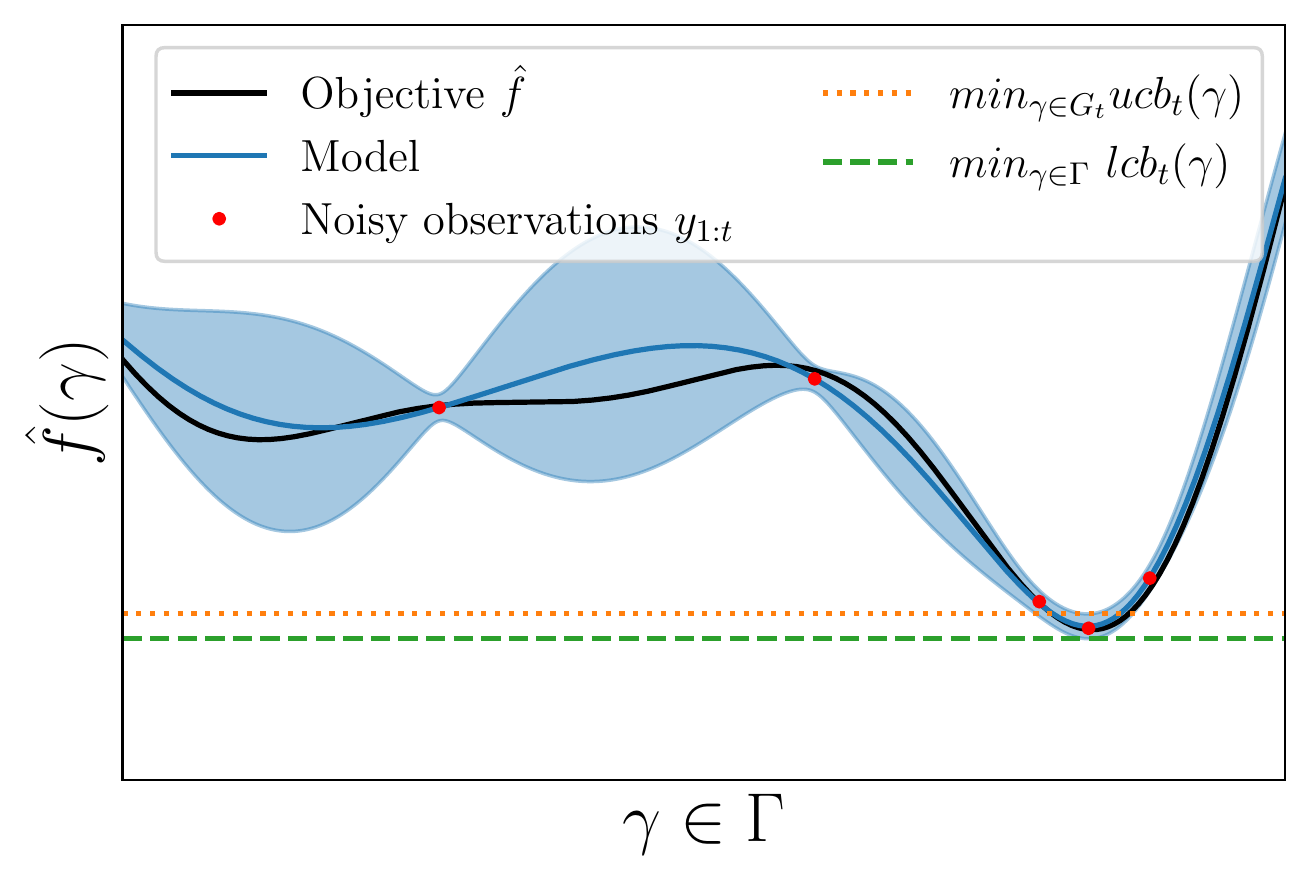}
 \includegraphics[width=0.49\textwidth]{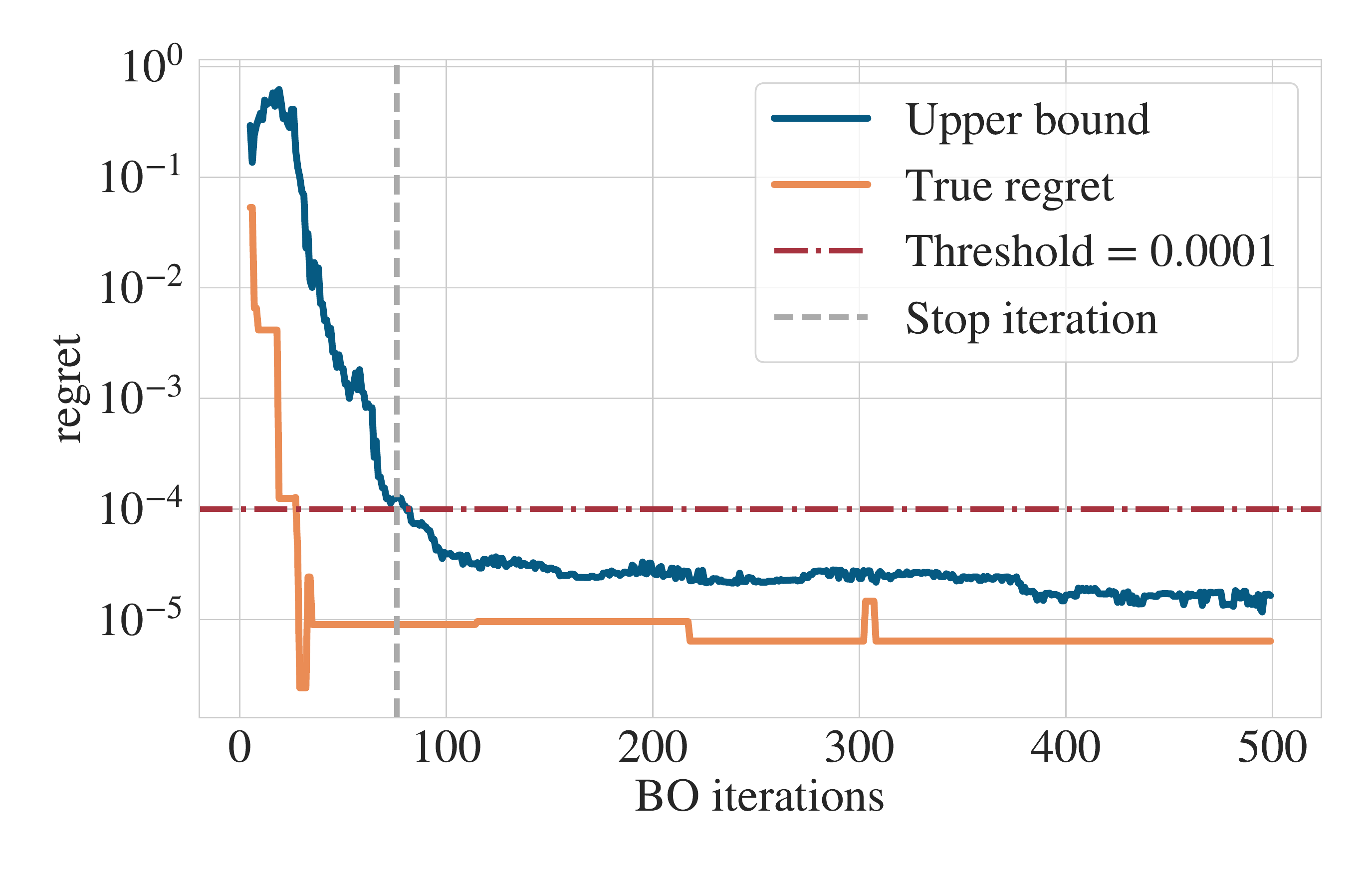}

 \caption{\textbf{Left}: Visualization of the upper bound for $\simpleregretestim_t$. The gap between green and orange lines is the estimate of the upper bound for $\simpleregretestim_t$. \textbf{Right}: Illustration of automated BO termination when tuning MLP on the \texttt{naval} dataset from HPO-Bench \citep{Klein2019Tabular} with the BORE optimizer \citep{pmlr-v139-tiao21a}. \looseness -1}
 \label{fig:bound-example}

\end{figure}

Consequently, we can bound the unknown  $\festim(\goptim_{t})$ and $\festim(\goptimD)$ that define the sub-optimality $\simpleregretestim_t$:\looseness-1
    \begin{align}
       \simpleregretestim_t = \festim(\goptim_{t}) - \festim(\goptimD) \leq  \underset{\gamma \in \Dt}{\min}\  \ucb_t(\gamma)  - \underset{\gamma \in \Gamma}{\min} \ \lcb_t(\gamma) \eqqcolon \bar{\simpleregret}_t, \label{eq:regret_bound}
\end{align} 
where the inequality for $\festim(\gamma^*_t)$ is due to the definition of the reporting rule $\gamma^*_t = \argmin_{\gamma \in \Dt} \festim(\gamma)$ over the evaluated points $\Dt = \{\gamma_1,\dots,\gamma_t\}$. We illustrate the idea with an example in \cref{fig:bound-example}.

\textbf{Termination threshold.} We showed how to control the optimization error via the (computable) regret upper bound $\bar{\simpleregret}_t$ above. We now explain when to stop BO, i.e., how to choose some threshold $\epsilon_{BO}$ and an iteration $T: \bar{r}_T \leq \epsilon_{BO}$. Following Proposition 1, we suggest setting $\epsilon_{BO}$ to be of similar magnitude as the statistical error $\epsilon_{st}$ of the empirical estimator (since smaller regret $\simpleregretestim_t$ is not beneficial when $\epsilon_{st}$ dominates). In case of cross-validation being used for HPO, one can estimate this statistical error $\epsilon_{st}$ and we further discuss how it can be done. 
% When cross-validation is not used, one could define an intuitive threshold in advance due to the usage of interpretable upper bound of simple regret. THE VERY SAME SENTENCE as in Section 3.1
%This means that whenever the (computable) $\bar{\simpleregret}_t$ is less than any threshold $\epsilon_{BO}$, BO is guaranteed to have identified an $\epsilon_{BO}$-optimal solution in terms of the empirical surrogate $\festim$. However, this threshold $\epsilon_{BO}$ still needs to be set appropriately.  \looseness-1

Cross-validation is the standard approach to compute an estimator $\festim(\gamma)$ of the population risk. %Add a reference; an old reference could to one of Efron's papers
The data $\D$ is partitioned into $k$ equal-sized sets $\D_1,\dots,\D_k$ %this is k-fold cross-validation
used for (a) training the model $\algo_{\gamma}(\cdot ; \D_{-i})$, where $\D_{-i}=\cup_{j\neq i}\D_i$ (i.e., training on all but the $i$-th fold), and (b) validating $\algo_{\gamma}(\cdot ; \D_i)$ on the $i$-th fold of the data. These two steps are repeated in a loop $k$ times, and then the average %CA: of what? 
over $k$ validation results is computed.
%, %. That yields 
%yielding the noisy BO evaluation $y = \festim(\gamma) + \varepsilon$ where the noise is due to the randomness in the training procedure or to epistemic uncertainty.%CA: this is inaccurate: it can be due the training procedure or to epistemic uncertainty   % done.

The statistical error $\epsilon_{st}$ of an estimator can be characterised in terms of its variance $\VarG(\gamma) = \E[(\festim(\gamma) - \E \festim(\gamma))^2]$ and bias $B(\gamma) = \E [\festim(\gamma)]- f(\gamma)$, where the latter can be neglected in case of cross-validation \citep{bayle2020crossvalidation}.  Though the variance $\VarG(\gamma)$ of the cross-validation estimate  is generally unknown, \cite{nadeau2003inference} propose an unbiased estimate for it. Specifically, for the sample variance (denoted as $\VarCV$) of $k$-fold cross-validation, a simple post-correction technique to estimate the  variance $\mathrm{Var}\festim(\gamma)$ is \looseness-1
\begin{align}
\VarG(\gamma) \approx \left( \frac{1}{k} + \frac{|\D_{i}|}{|\D_{-i}|} \right) \VarCV(\gamma), \label{eq:variance_estimation}
\end{align}
\looseness-1 where $|\D_{i}|,  |\D_{-i}|$ are the set sizes. For example, in the case of 10-fold cross-validation we have $\VarG(\gamma) \approx 0.21\VarCV(\gamma)$. We are now ready to propose our termination condition in the following.\looseness-1

\textbf{Termination condition for BO.} \textit{Consider the setup of Proposition 1 where $\hat f(\cdot)$ is a cross-validation-based estimator being iteratively minimized by BO. Let $\gamma_t^*$ the solution reported in round $t$,
%Let $y_t = \festim(\gamma_t) + \varepsilon_t$ be a noisy evaluation at the input $\gamma_t$ and $\gamma_t^* = \underset{\scriptstyle{\gamma_i \in \{\gamma_1,\dots,\gamma_t\}}} {\argmin}\ y_i$ denote the best observed input by BO iteration $t$.  Let 
and 
$\bar{\simpleregret}_t$ defined in \cref{eq:regret_bound} be the simple regret bound computed at each iteration $t$. Let the variance $\VarG(\gamma_t^*)$ of the estimator $\hat f(\cdot)$ be approximated according to  \cref{eq:variance_estimation}. Then, BO is terminated once:\looseness-1} 
\begin{equation}
    \bar{\simpleregret}_t < \sqrt{\VarG(\gamma_t^*)}.
\end{equation}

Intuitively, the termination is triggered once the maximum plausible improvement becomes less than the standard deviation of the estimate. This variance-based termination condition adapts to different algorithms or datasets and its computation comes with negligible computational cost on top of cross-validation. The pseudo-code for the criterion is summarised in \cref{app:algo}. If cross-validation cannot be used or is computationally prohibitive, the user can define the right-hand side of the termination condition. In this case, the upper bound on the left-hand side still has an intuitive interpretation: the user can set the threshold based on their desired solution accuracy. This case is demonstrated in \cref{fig:bound-example}, with an example of automatic termination for tuning an MLP. \looseness-1

\section{Experiments}
\label{sec:exp}

The main challenge of any termination criterion for HPO is to balance between reducing runtime and performance degradation.
We thus study in experiments how the speed-up gained from different termination criterions affects the final test performance. To this end, we define two new metrics that account for the trade-off between resources saved and performance drop and provide a list of reasonable baselines. The code to reproduce the experimental results is publicly available.\footnote[1]{\url{https://github.com/amazon-research/bo-early-stopping}} \\

\textbf{Baselines.} %HPO early stopping is overlooked in the research community, we collect several at our best in the following. 
Since automatic BO terminating is a rather under-explored topic, we consider the following baselines that are, to the best of our knowledge, the only ones directly related to our method: \looseness-1
\begin{itemize}
\item N\"aive convergence test controlled by a parameter $i$ (referred as Conv-$i$): stopping BO if the \textit{best} observed validation metric remains unchanged for $i$ consecutive iterations. It is challenging to define a suitable $i$ suitable across different benchmarks since $i$ is a fix, predetermined choice, that does not take the observed data into account (in contrast to our method that refines the regret estimation).  We consider common in practice values $i = \{10, 30, 50\}$ and study other values in Appendix.  \looseness-1

\item Threshold for Expected improvement \citep[EI;][]{Nguyen2017regret}: stopping BO once EI drops below a pre-defined threshold. 
Choosing a threshold crucially depends on a problem at hand, e.g., values studied by the original paper result in too aggressive stopping across a range of our experiments. We thus extend it with a more finer grained grid resulting into $\{10^{-9}, 10^{-13}, 10^{-17}\}$.  \looseness-1
% of $\{10^{-9}, 10^{-13}, 10^{-17}\}$. 

\item Threshold for Probability of improvement \citep[PI;][]{lorenz2016stopping}: stopping BO once PI drops below a pre-defined threshold. Similar to EI baseline, we tune the threshold and use  $\{10^{-5}, 10^{-9}, 10^{-13}\}$.  \looseness-1

% $\{10^{-4}, 10^{-8}, 10^{-12}\}$

% Empirically, we observe that all thresholds lead to significant speeds up but, at the same time, to a severe degradation of the final solution quality (see \cref{app:tabres} for full results).

%CA: I guess it is too late, but it might have been useful to consider a more finer grained grid; maybe something to consider in anticipation to the rebuttal.

\end{itemize}
 %This method mimics the early stopping during algorithm training with two notable differences: First, we only track the validation metrics of the incumbent instead of the suggested hyperparameters at every iteration because the latter may underperform due to the explorative nature in BO. Second, defining a threshold is not necessary as the incumbent may stay the same for many iterations and then suddenly change, as shown in \cref{fig:test_jumps}.  and BO budget $T = 200$.

\textbf{Metrics.}
We measure the effectiveness of a termination criterion via two metrics quantifying (i) the change in test error on a held-out dataset and (ii) the time saved. Given a budget of $T$ BO iterations, we compare the test error  $y_T$ after $T$ iterations to the test error $y_{es}$ after early stopping is triggered. For each experiment, we compute the \textit{relative test error change} RYC as
\begin{equation}
    \text{RYC} =  \frac{y_T - y_{es}}{\max(y_T, y_{es})}. \label{eq:ryc}
\end{equation}
This allows us to aggregate the results over different algorithms and datasets, as RYC $\in [-1, 1]$ and can be interpreted as follows: A positive RYC represents an improvement in the test error when applying early stopping, while a negative RYC indicates the opposite. Similarly, let the total training time for a predefined budget $T$ be $t_T$ and the total training time when early stopping is triggered be $t_{es}$. Then the \textit{relative time change} RTC  is defined as
\begin{equation}
\text{RTC} =  \frac{t_T - t_{es}}{t_T} \label{eq:rtc}
\end{equation}
indicates a reduction in total training time, RTC $\in [0, 1]$. While reducing training time is desirable, it should be noted that this can be achieved through any simple stopping criterion (e.g., consider interrupting HPO with a fixed probability after every iteration). In other words, the RTC is not a meaningful metric when decoupled from the RYC and, thus, both need to be considered in tandem. \\

\begin{figure}[h!]
\centering
 
\includegraphics[width=0.49\textwidth]{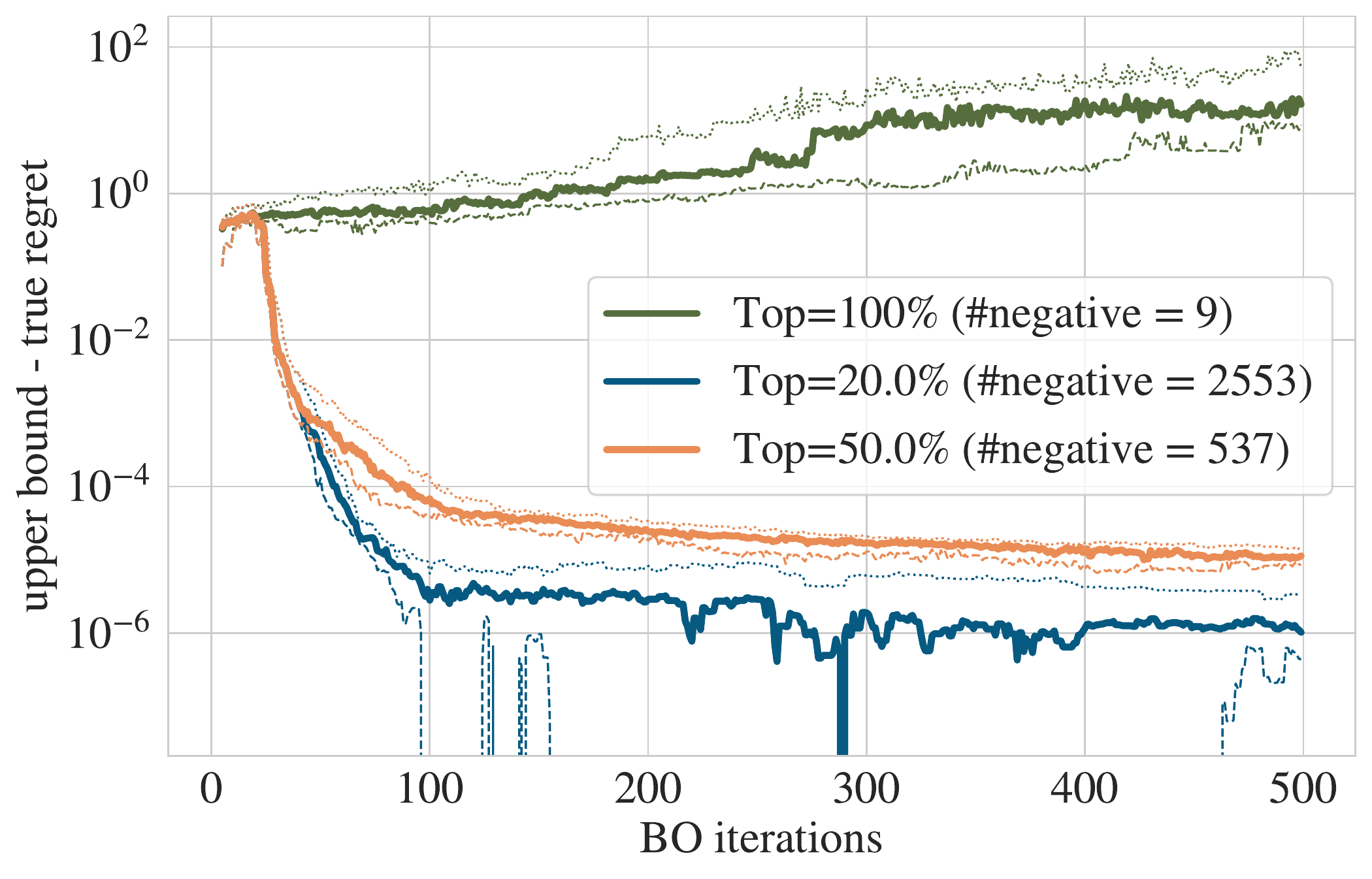}
\includegraphics[width=0.49\textwidth]{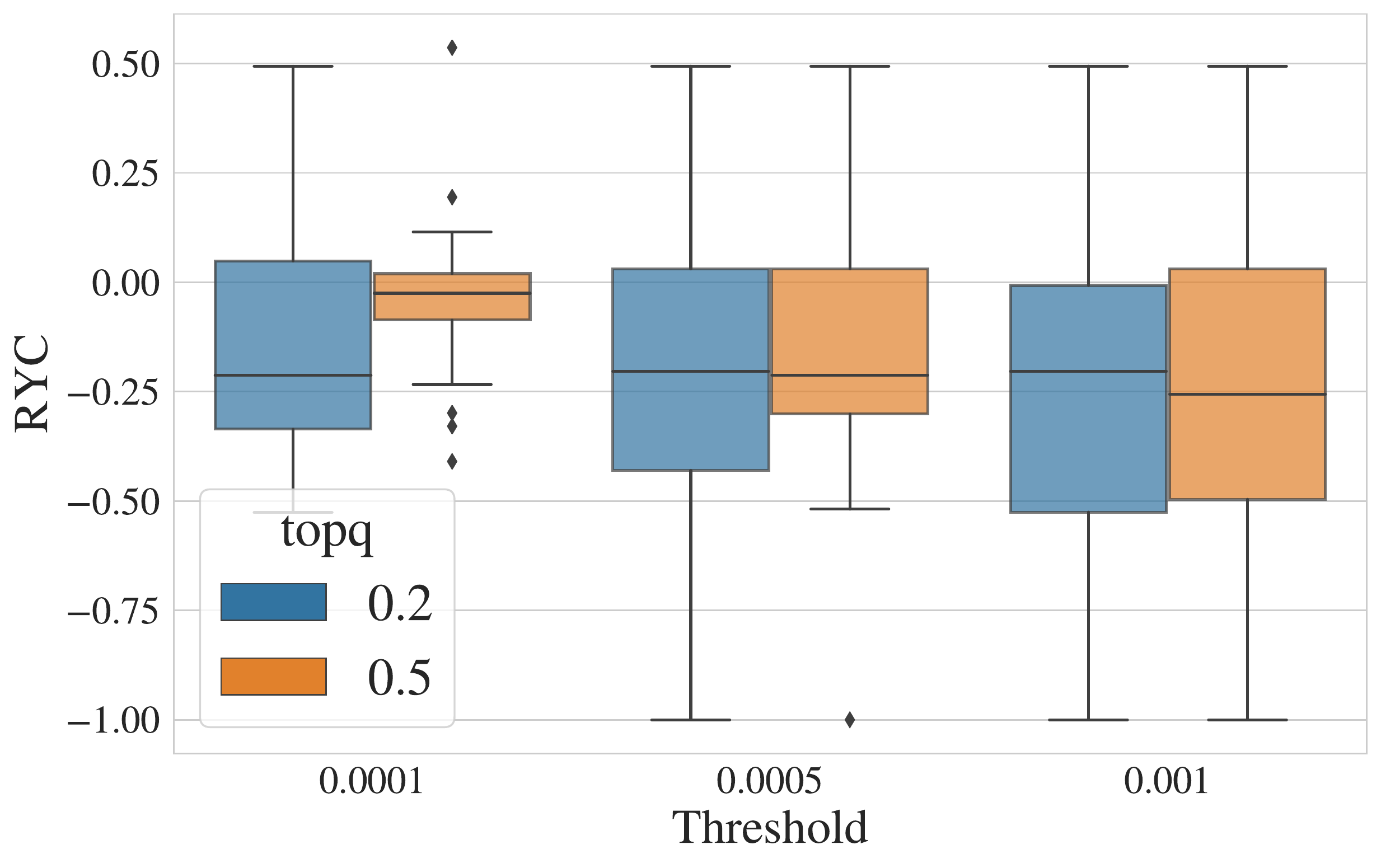}
  \caption{The upper bound estimation quality is affected by the set of hyperparameters evaluations used in the surrogate model training. \textbf{Left}: Bound quality for using all, top 50\% and top 20\%  hyperparameter evaluations, measured by the difference between upper bound and true regret.  Solid line represents the median over 50 replicates, the dashed is 20'th quantile and dotted line is 80'th quantile. The legend also shows the number of negative differences (the upper bound is smaller than the true regret).  \textbf{Right}: Box plots of RYC scores when using the top 50\% and top 20\% hyperparameter evaluations under common thresholds.}
  \label{fig:topq}
\end{figure}

\textbf{Selecting the data for the bound estimate.} Since we are only interested in the upper bound of the simple regret, we conjecture that using only the top performing hyperparameter evaluations may improve the estimation quality. To validate this, we use results of BORE \citep{pmlr-v139-tiao21a} %CA: this might require a bit more explanation; not everyone know BORE
on the \texttt{naval} dataset from HPO-Bench \citep{Klein2019Tabular} where we can quantify the true regret (see Section~\ref{ssec:nn}). We compute the upper bound by \cref{eq:regret_bound} using three options: 100\%, top 50\% or top 20\% of the hyperparameters evaluated so far and measure the distance to the true regret (see \cref{fig:topq}). From \cref{fig:topq}, fitting a surrogate model with all the hyperparameter evaluations poses a challenge for estimating the upper bound of the regret, which is aligned with recent findings on more efficient BO with local probabilistic model, especially for high-dimensional problems \citep{NEURIPS2019_6c990b7a}. Using the top 20\% evaluations gives the best upper bound estimation quality in the median, at the cost of the most under-estimations of the true regret (2553). Our method would stop too early due to the under-estimation, thus negatively impacting the RYC score, as shown in \cref{fig:topq}. \textit{As a result, we use the top 50\% hyperparameters evaluations for the upper bound estimation throughout this paper}.  

\subsection{BO for hyperparameter tuning with cross-validation}%CA: there might be a hyphenation inconsistency.
\label{ssec:cv}

We tune XGBoost (XGB; 9 hyperparameters) and Random Forest (RF; 3 hyperparameters) on 19 small tabular datasets, where we optimize error rate for classification and rooted mean square error for regression, computed via 10-fold cross validation. \looseness-1

\textbf{Methods setup.} We use a Mat{\'e}rn 5/2 kernel for the GP and its hyperparameters are found based on a type II maximum likelihood estimation (see \cref{sec:setting} for more details). The termination is triggered only after the first 20 iterations to ensure a robust fit of GP. We use \cref{eq:variance_estimation} to compute our stopping threshold, i.e., $\VarG(\gamma) \approx 0.21\VarCV(\gamma)$. We additionally use empirical scaling $0.5\VarCV(\gamma)$ to study the affect of varying the magnitude.  \looseness-1
%The only hyperparameter involved in our method is $\beta_t$ We use Theorem 1 in \cite{srinivas10} to set $\beta_t$ with a caveat: we take the number of hyperparameters in the search space as the size of the input space. We then further scale it down by a factor of 5 as defined in the experiments in \cite{srinivas10}. This computation of $\beta_t$ is fixed throughout our experiments. %CA: it is actually not clear why this is ok -- would it be possible to elaborate? A different strategy would have been to run this for multiple valued of beta,.

\begin{figure}[h!]
\centering
 \subfloat{
   \includegraphics[width=0.49\textwidth]{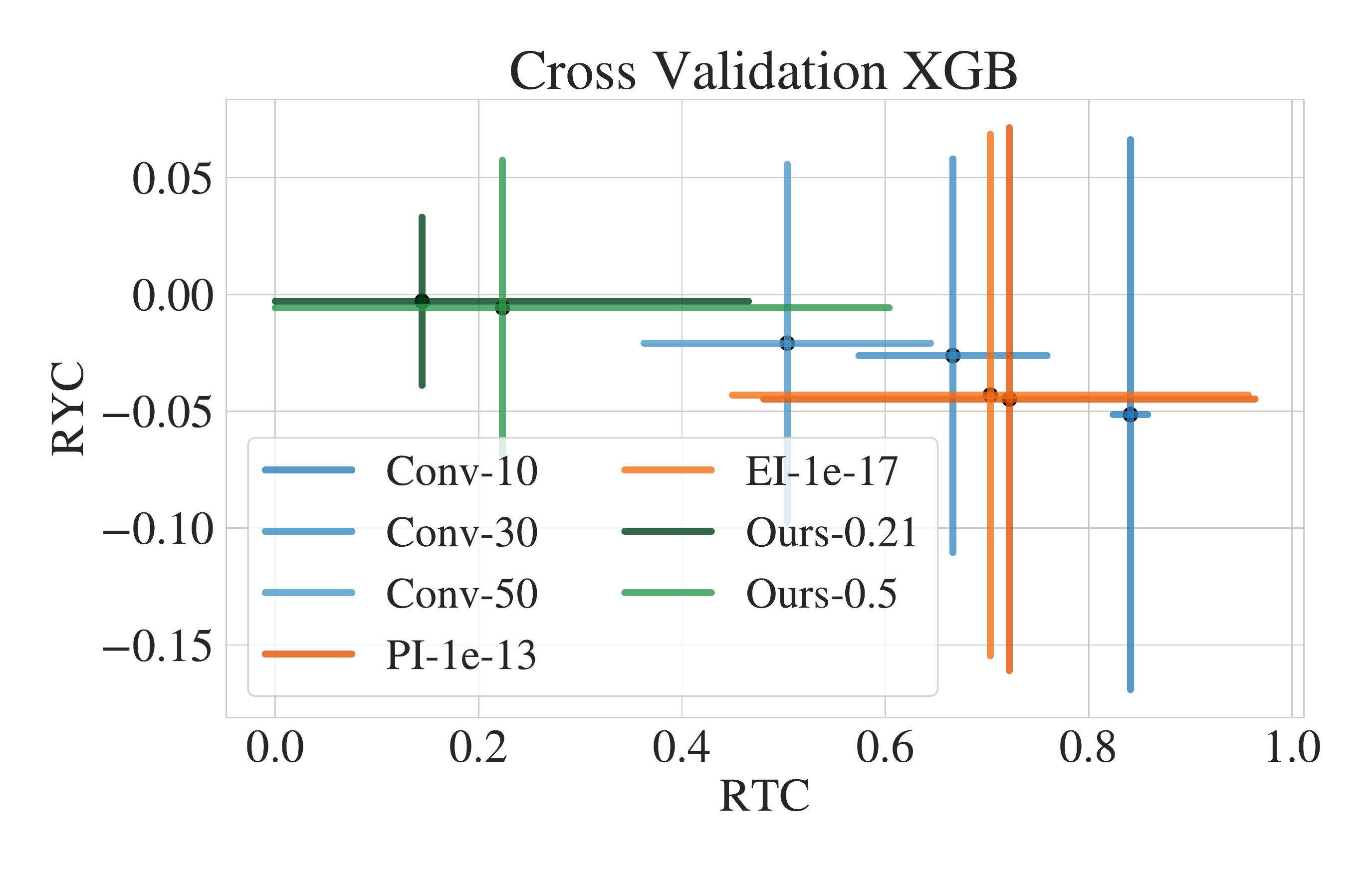}
  }
%   \hspace{0.05mm}
   \subfloat{
   \includegraphics[width=0.49\textwidth]{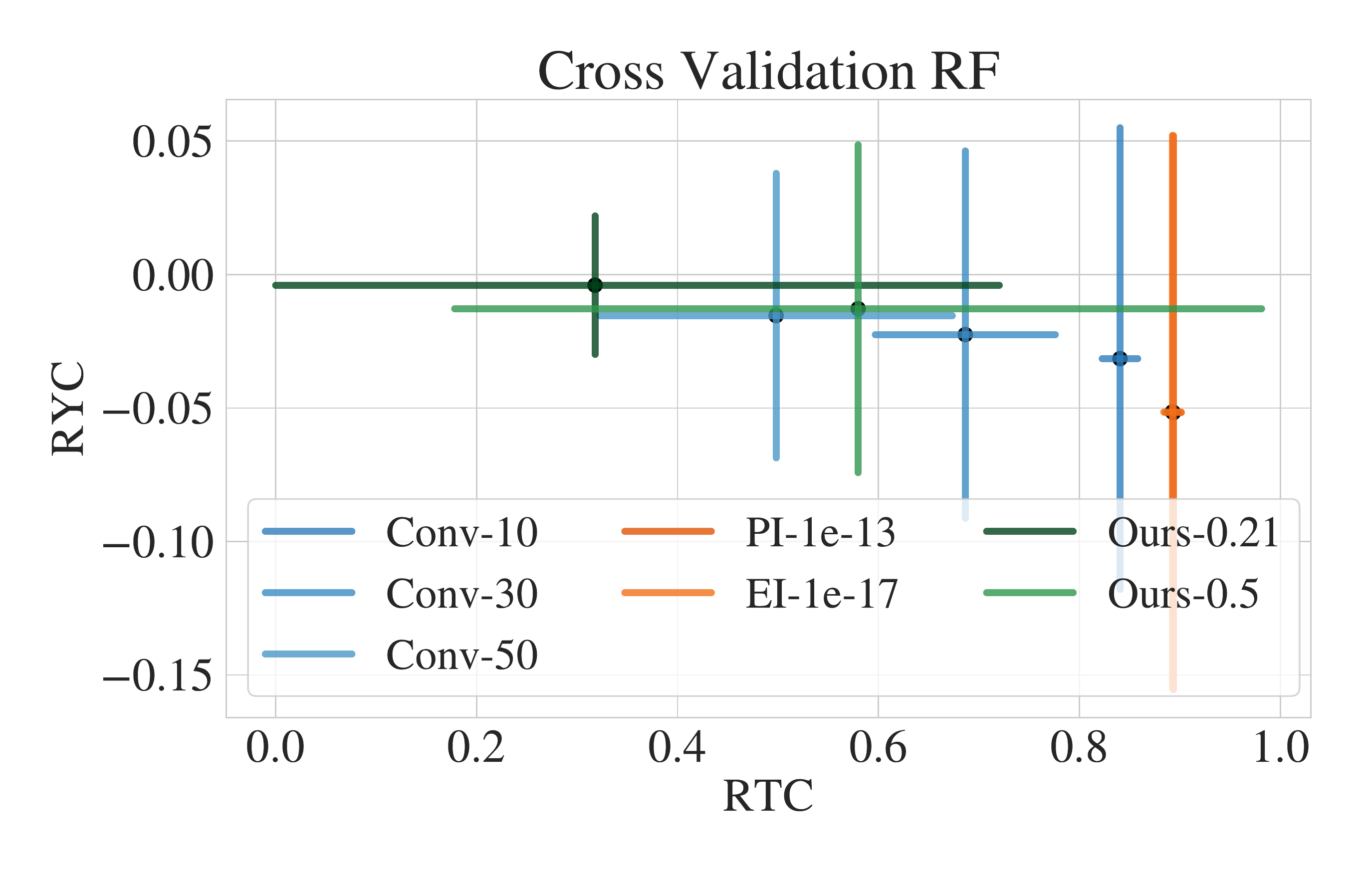}
 }
 \caption{The mean and standard deviation of RYC and RTC scores for the compared automatic termination methods when using cross validation in the hyperparameter evaluation when tuning XGB (left) and RF (right). The mean value is shown as the large dot and the standard deviation is shown as an error bar in both dimensions. \label{fig:cv}}
\end{figure}

 \textbf{Results.}  We present the RYC-RTC results aggregated across the datasets in \cref{fig:cv}, \cref{tab:cv} and \cref{fig:bo-baselnes-conv,fig:bo-baselnes} (\cref{sec:setting}).  The main take-away from \cref{fig:cv} is as follows: more aggressive early stopping might indeed speed up but it leads to worse test performance, both in terms of average and standard deviation over the datasets. In contrast, the desirable behavior is the trade-off between RYC and RTC, where lower RYC error bars are prioritized over lower RTC error bars. In other words, the methods that adaptively stops BO, i.e., stops when it necessary and does not stop when it is not, is preferable. \cref{fig:cv} indicates that our method successfully maintains a high solution quality across a wide range of scenarios (lower RYC variance) via adapting the termination to the particular propblem (higher variance RTC). The results also show an anticipated RYC-RTC trade-off for $i$ in Conv-$i$ and thresholds in EI and PI baselines, where the solution quality improves as the thresholds increases and consequently the speed-up drops. The average RYC results in \cref{tab:cv} however are dominated by our method.  \looseness-1

% Furthermore, the standard deviations of the RYC and RTC scores also provide us with interesting insights. The RYC variances of our method are usually smaller than the baselines ones, indicating that we successfully maintain a high solution quality across a wide range of scenarios. On the other hand, the RTC variances of our method are usually higher than the baselines, which highlights that our method adapts to different scenarios rather than stopping BO at similar iterations. 
%These results show that our approach compares favourably to baselines by achieving competitive RYC scores while stopping early. %This is a much more challenging task than purely speeding up HPO, which can be easily achieved through trivial stopping strategies (e.g., by decreasing $i$).

%The LM results suggest that stopping early (by PI or EI based baselines) is beneficial. 
%To further demonstrate that our method is the most adaptive comparing to the baselines, we plot the histogram of iterations that BO starting to overfit (the iteration that test error starting to increase) in all the experiments, as well as the iterations that are being stopped by different methods in \cref{fig:iter}. Ideally, we expect when overfitting happen, an early stopping method will be triggered and otherwise being silent. From \cref{fig:iter}, it is clear that other baselines  are triggered mostly in early BO stage while our method being the only one taking action accordingly when BO does not overfit or overfit in later stage.

%\begin{figure}[ht]
%\centering
%\includegraphics[width=\columnwidth]{imgs/overfit_iter_hist.eps}
%\caption{Histogram of iterations that BO starts to overfit (test error starts to increase), as well as the iterations that are being stopped by different BO early stopping methods. Our method being the only one taking action accordingly when BO does not overfit or overfit in later stage.}
%\label{fig:iter}
%\end{figure}

\subsection{Neural Hyperparameter and Architecture Search}
\label{ssec:nn}
%\aaron{So far we have more qualitative results, I think we also have some quantative results, i.e having a table as we had before . I will generate more results for the NASbenchmarks with different optimiziers}
In these experiments we study out termination criterion beyond the BO scope, showing its main advantage of being applicable for any iterative HPO method. In addition, here we also show how to use our method if cross-validation is unavailable. To demonstrate this, we apply it to several state-of-the-art methods: TPE~\citep{bergstra2011algorithms}, BORE~\citep{pmlr-v139-tiao21a}, GP-BO~\citep{snoek12} as well as random search (RS)~\citep{JMLR:v13:bergstra12a}. 

\textbf{Benchmarks.} We consider two popular tabular benchmark suites: NAS-HPO-Bench \citep{Klein2019Tabular}, which mimics the hyperparameter and neural architecture search of multi-layer perceptrons on tabular regression datasets, and NAS-Bench-201 \citep{dong2020nasbench201} for neural architecture search on image classification datasets. Notice that for NAS-Bench-201, we used \textit{validation} metrics to compute RYC instead of test metrics, thus, no positive RYC scores are observed. For a detailed description of these benchmarks we refer to the original paper. 

\textbf{Methods setup.}  We consider the following thresholds on the final regret $\{0.0001, 0.001, 0.01\}$, corresponding to a loss of performance of $0.01\%$, $0.1\%$ and $1\%$, respectively. Note that, it is easier to set the threshold for our method because it is a threshold on the regret in the metric space that users aim to optimize and it gives more explicit control of this trade-off, thus making it more interpretable. For each method and dataset, we perform 50 independent runs with a different seed.

\textbf{Results.} \cref{fig:hpo-bench} and \cref{fig:nas-bench} show results for BORE and the rest can be found in the Appendix. While \emph{no method} Pareto dominates the others, our termination criterion shows a similar trend as in \cref{ssec:cv} by prioritising accuracy over speed. Users choose the threshold based on their preference regarding the speed-accuracy trade-off, i.e, a higher threshold saves more wall-clock time but potentially leads to a higher drop in performance. %However, our criterion benefits from an intuitive and interpretable threshold. 
We further show the distribution of true regrets at the stopping iteration triggered by our method with the considered thresholds on HPO-Bench in \cref{fig:nas-bench}. \looseness-1

From \cref{fig:nas-bench}, with a high threshold of $0.01$, all the experiments (4 datasets with 50 replicates) are early stopped by our method and 41 (20\%) experiments end up with true regret being higher than the threshold.  
With a low threshold of 0.0001, 112 experiments are stopped and 12 (10.7\%) experiments end up with true regret above the threshold. In short, our method achieves 80\% to 90\% success rate where the true regret is within the user-defined tolerance.

% \begin{figure}[h!]
% \centering

%  \subfloat{
%   \includegraphics[width=0.45\textwidth]{plots/bore_hpobench_naval_scatter.pdf}
%   }
%   \hspace{1mm}
%  \subfloat{
%   \includegraphics[width=0.45\textwidth]{plots/bore_hpobench_parkinsons_scatter.pdf}
%   }    	\\
%   \subfloat{
%   \includegraphics[width=0.45\textwidth]{plots/bore_hpobench_protein_scatter.pdf}
%   }
%   \hspace{1mm}
%  \subfloat{
%   \includegraphics[width=0.45\textwidth]{plots/bore_hpobench_slice_scatter.pdf}
%   }  
%   \caption{The mean and standard deviation of RYC and RTC scores for considered automatic termination methods for HPO-Bench datasets. The mean value is shown in the large dot and the standard deviation is shown as an error bar in both dimensions.\label{fig:hpo-bench}}
% \end{figure}

\begin{figure}[h!]
\centering
  \includegraphics[width=0.49\textwidth]{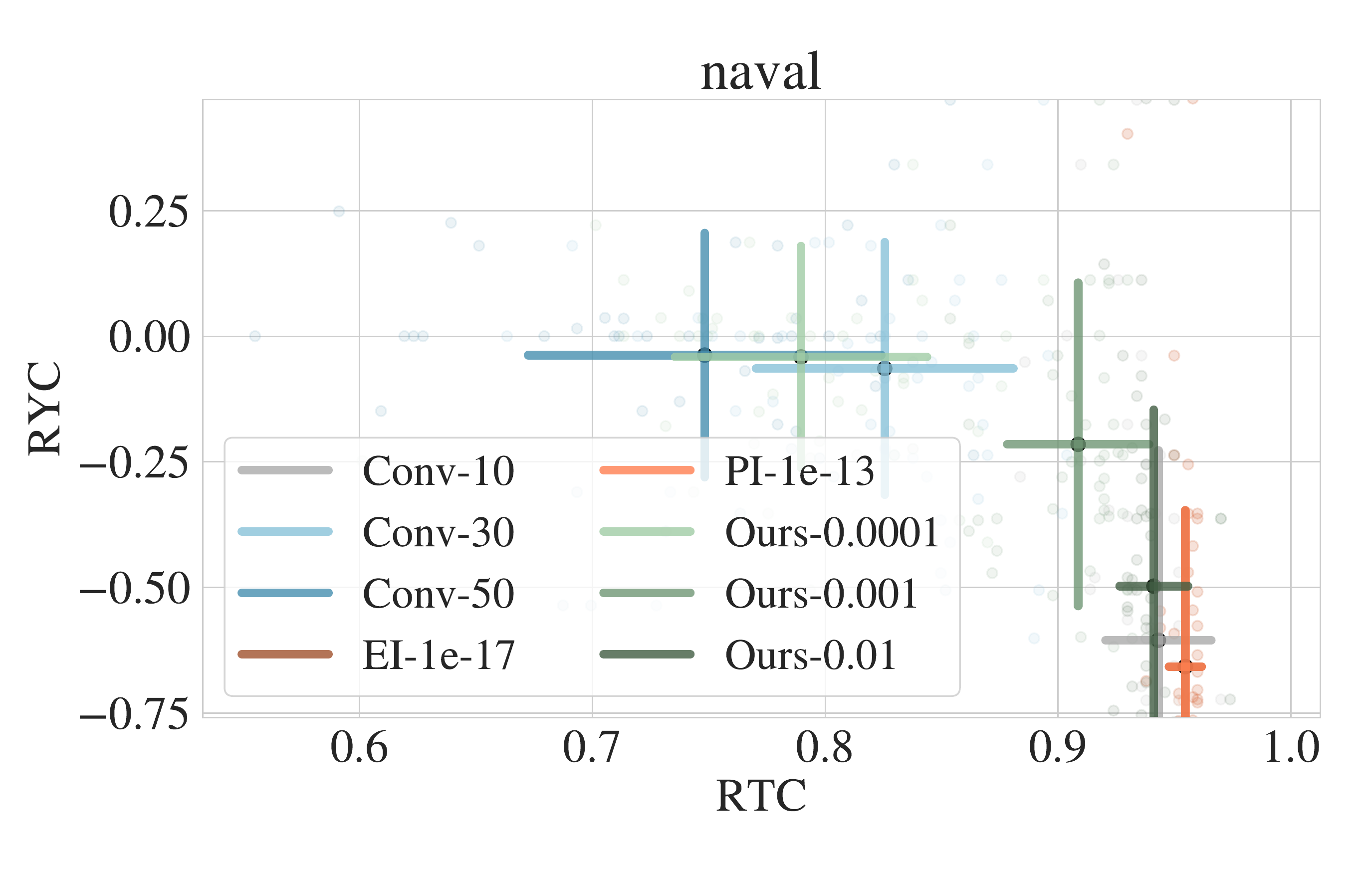}
  \includegraphics[width=0.49\textwidth]{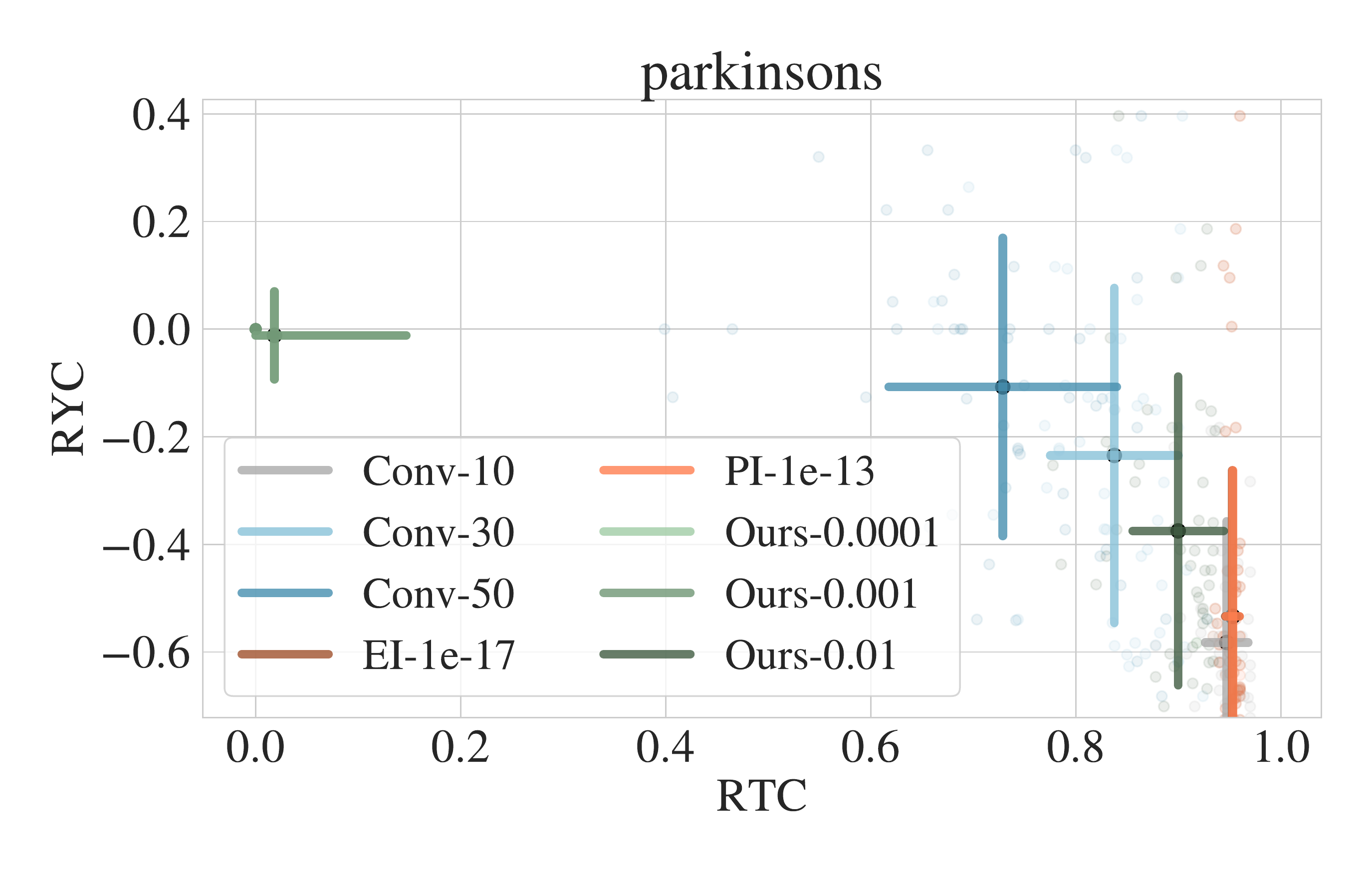} \\
  \includegraphics[width=0.49\textwidth]{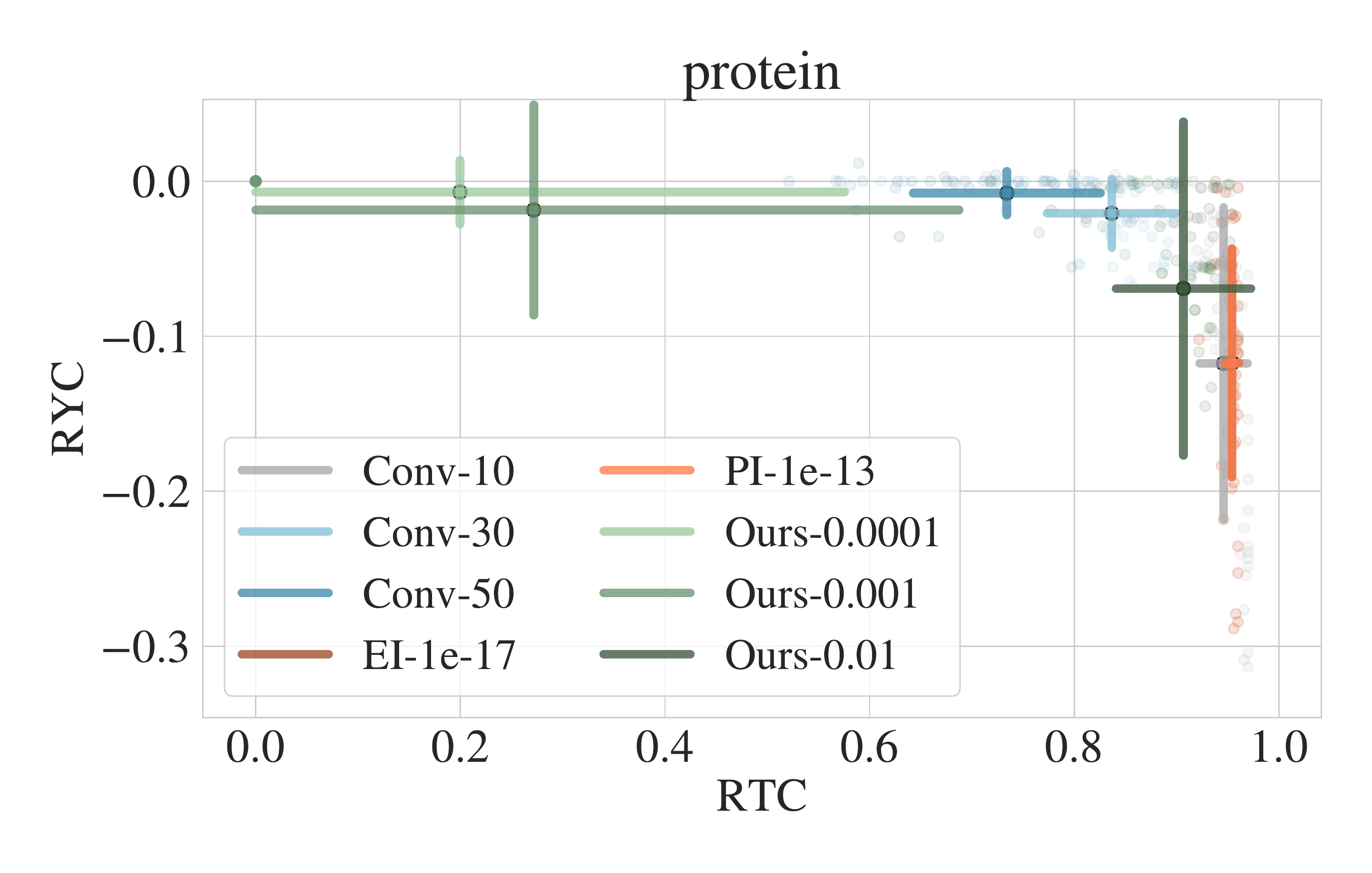}
  \includegraphics[width=0.49\textwidth]{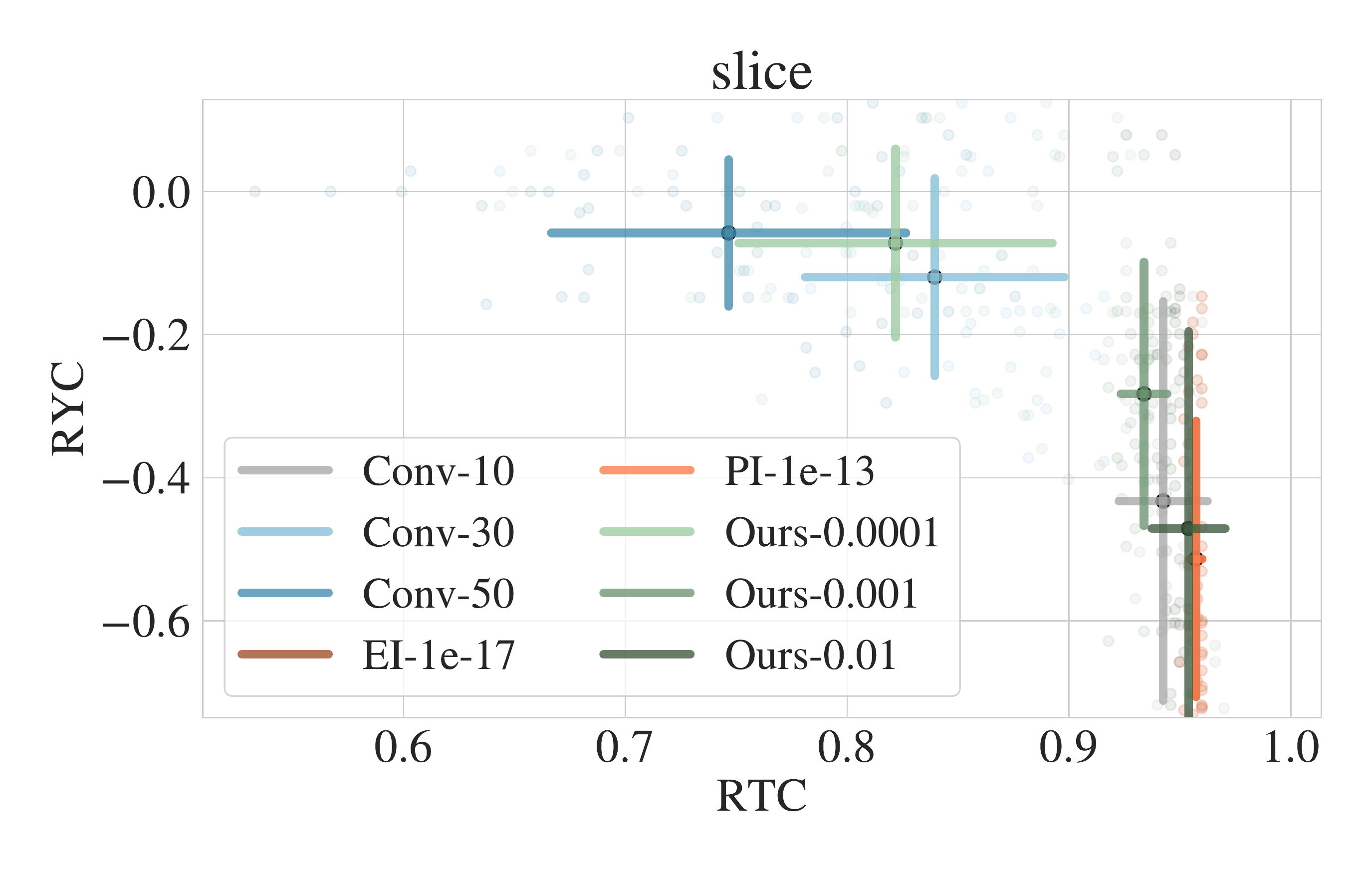}
  \caption{Mean (large dot) and standard deviation (error bar) of RYC and RTC scores for all methods for HPO-Bench datasets. \label{fig:hpo-bench}}
\end{figure}

\begin{figure}[h!]
\centering
 \includegraphics[width=0.49\textwidth]{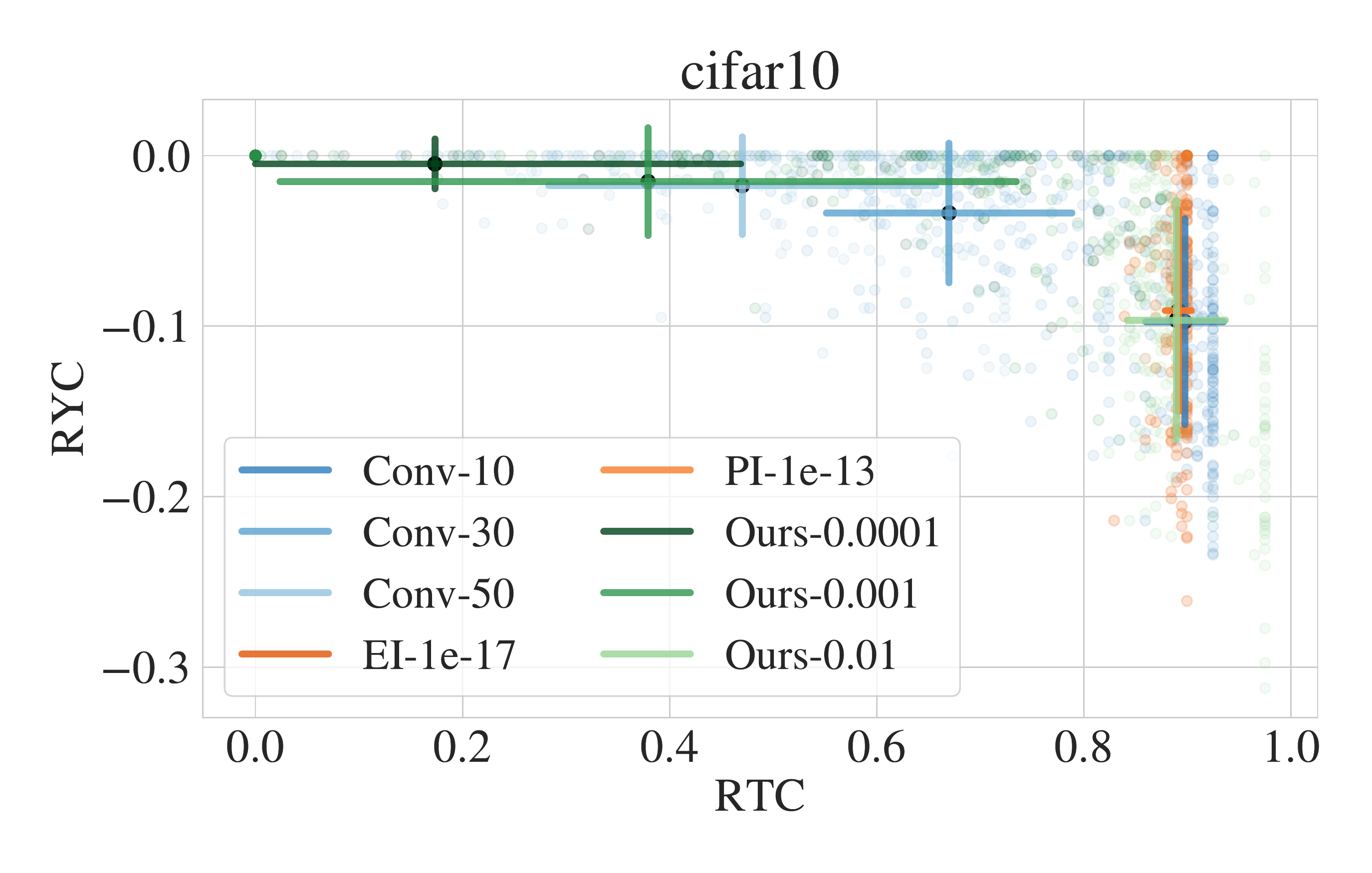}
 \includegraphics[width=0.49\textwidth]{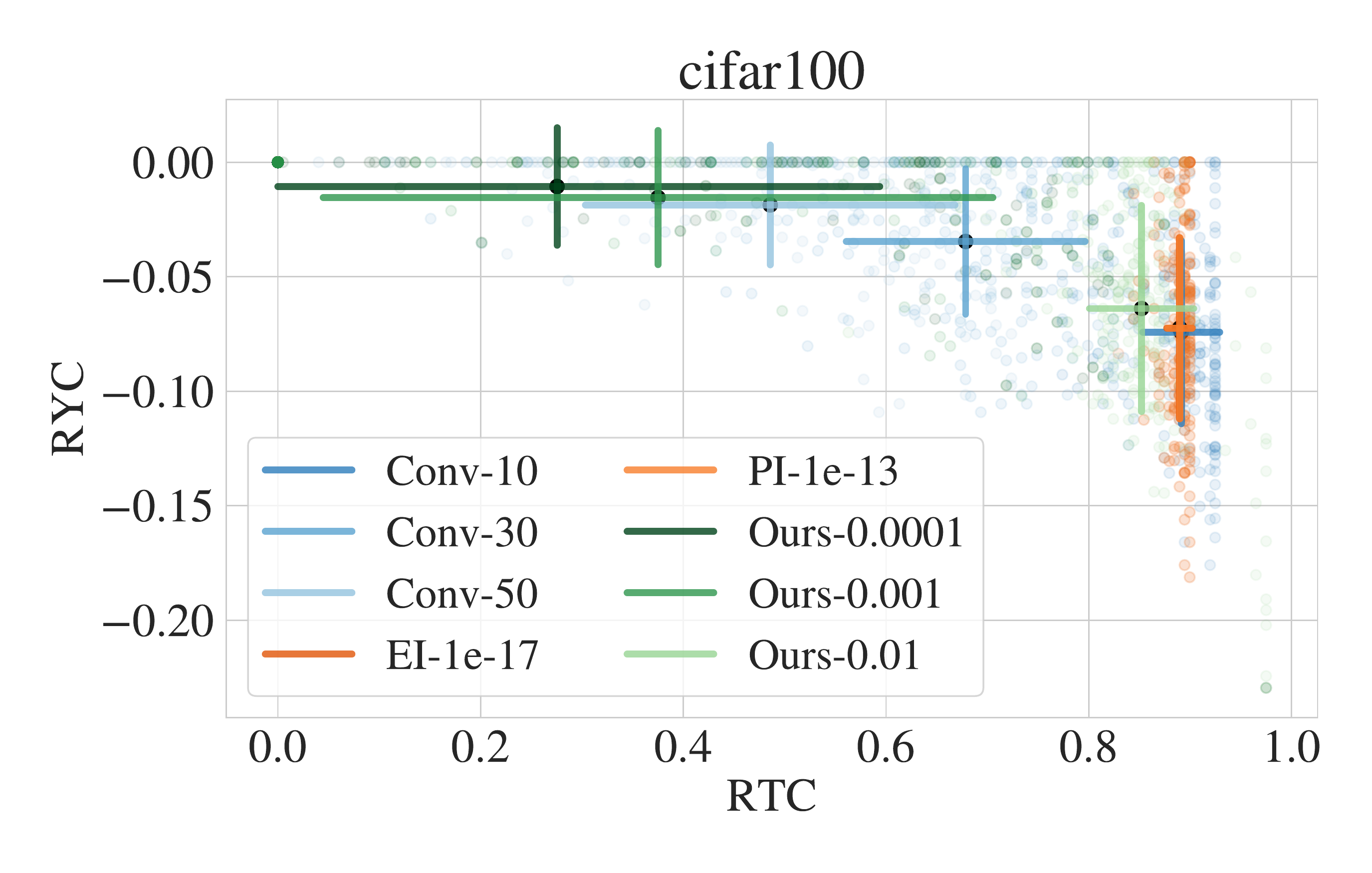} \\
 \includegraphics[width=0.49\textwidth]{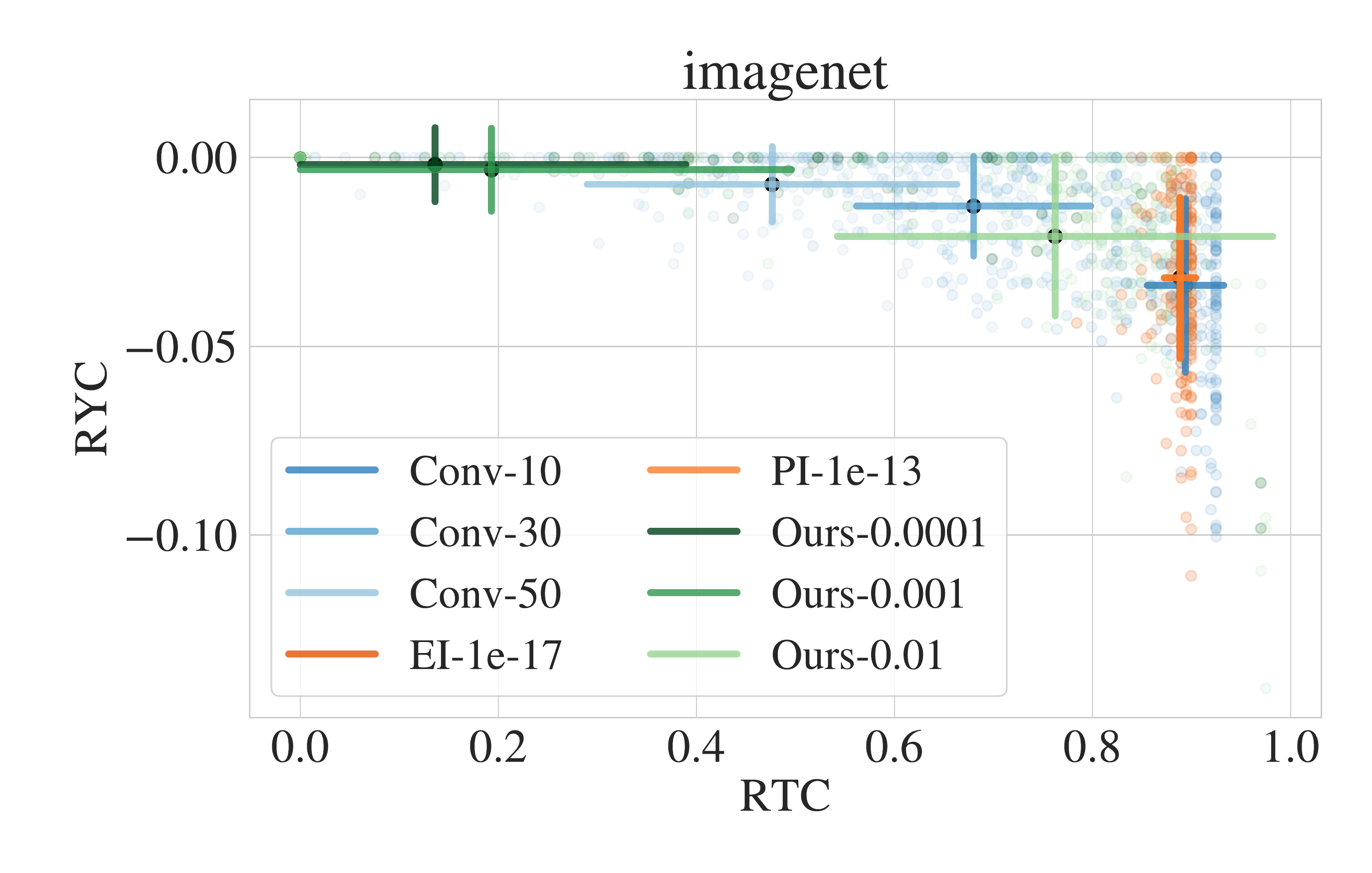}
 \includegraphics[width=0.49\textwidth]{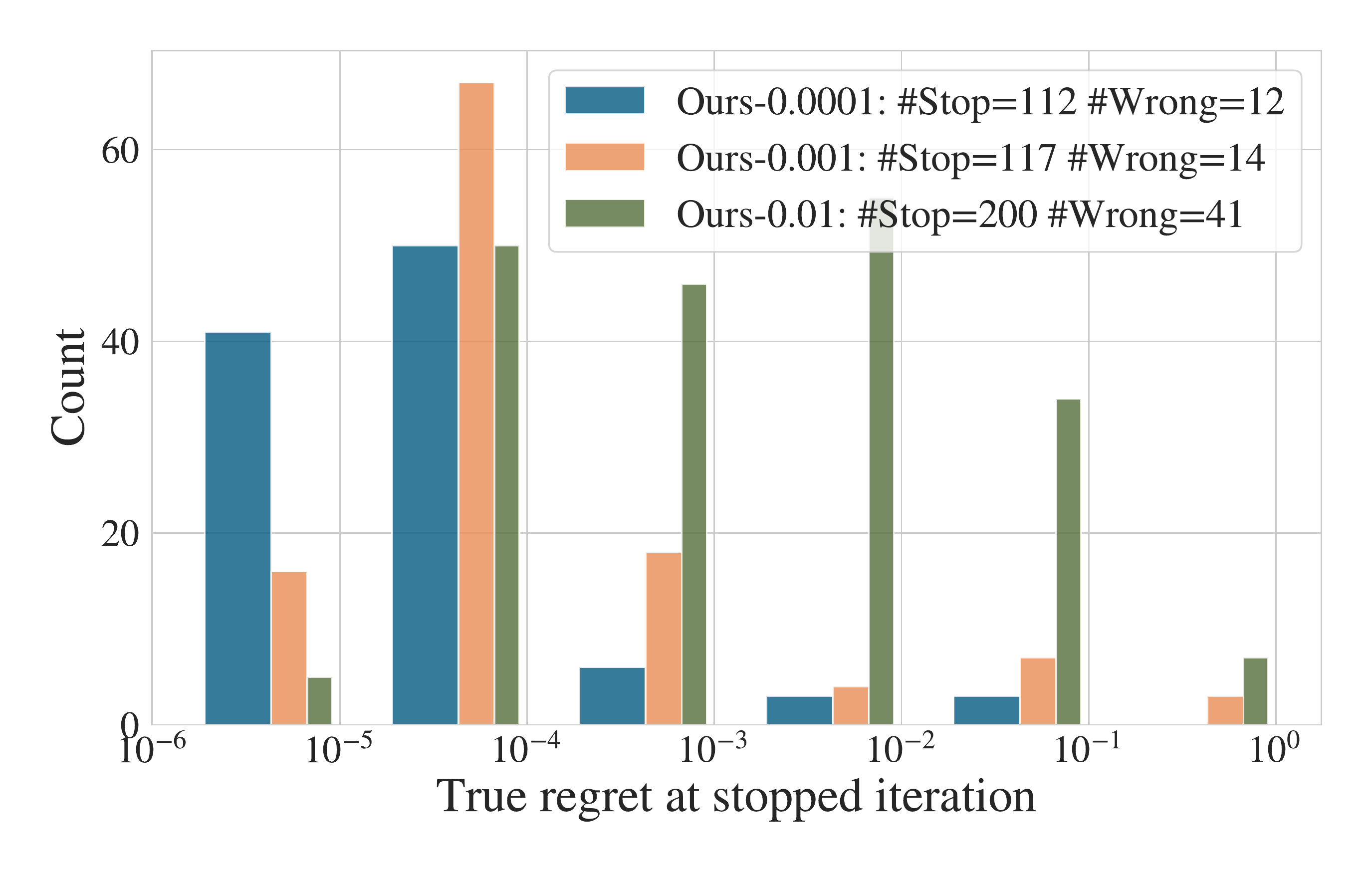}
 \caption{First three figures from the left show the mean and standard deviation of RYC and RTC scores for all methods on NAS-Bench-201. Since \textit{validation metrics} are used in these experiments,  no positive RYC scores are observed. The mean value is shown in the big dot and the standard deviation is shown as error bar in both dimensions. The Figure on the right shows a distribution of true regrets at the stopping iteration triggered by our method with different thresholds for HPO-Bench. %The number of stopped experiments and the number of ``wrong''  cases where the true regret is larger than the threshold are shown in the legend. 
 \label{fig:nas-bench}}
\end{figure}

%\subsubsection{HPO-Bench with other optimizers}

%We further compare our early stopping method to the convergence check baselines on two more optimizers besides BORE: Random Search (RS) and BO with TPE \cite{bergstra2011algorithms}. 
For every method we aggregated the scores over datasets with other HPO optimizers in \cref{fig:hpo-bench-opt} (see \cref{app:tabres}). We can see that the speed up of the convergence check baseline is affected very mildly by the optimizers while the RYC scores largely depend on the optimiser: RYC scores with random search are worse than with BORE. In contrast, the RYC scores for our termination criterion are similar across optimisers, especially for smaller thresholds. On the other hand, the speed up for a given threshold tends to vary. This can be explained by the difference of the optimizer's performance, for examples random search is not as efficient as BORE, and hence the regret is  mostly above the stopping threshold. In summary, while convergence check baselines are by design robust in terms of time saved, our method is more robust in terms of maintaining the solution quality.

\subsection{Overfitting in BO for Hyperparameter Optimization}

Proposition 1 emphasises an important problem of BO-based HPO: while focusing (and minimizing) the validation error, we cannot fully reduce the discrepancy between the validation and test errors. 
%Thus, it might be a case that the test error grows leading to overfitting. 
Empirically, we show this might happen when correlation between the test and validation errors is low, thus improvement in validation performance does not lead to the better test results. A particular example of such low correlation in the \textit{small} error region is presented in \cref{fig:corr}  (\cref{app:tabres}) when tuning XGB and Random Forest on tst-census dataset.

%While the validation errors of hyperparameter configurations are effectively reduced by HPO optimizers, the test errors may grow, \textit{i.e.} the HPO process can also overfit. We conjecture it originates from the statistical error $||\festim-f||_\infty \leq\epsilon_{st}$ in Proposition 1. We show the correlation between validation and test errors in the \textit{low} error region are indeed not well correlated through an example of tuning XGB and RF on \texttt{tst-census} dataset in \cref{fig:corr} in the \cref{sec:tabres}.

In our experiments, a positive RYC score is an indicator of overfitting, showing that the test error at the terminated iteration is lower than the test error in the final round. We observe positive RYC scores in both \cref{fig:cv} and \cref{fig:hpo-bench}, one with cross-validation on small datasets and one with medium-sized datasets. Hence, we %attempt to call for 
would like raise attention to the possible overfitting issue that occurs in HPO for which our method can be used as a plugin to mitigate overfitting.

\section{Conclusion}
\label{sec:disc}

Despite the usefulness of hyperparameter optimizations (HPO), setting a budget in advance remains a challenging problem. In this work, we propose an automatic termination criterion that can be plugged into many common HPO methods. The criterion uses an intuitive and interpretable upper bound of simple regret, allowing users explicitly control the accuracy loss. In addition, when cross-validation is used in the evaluations of hyperparameters, we propose to use an analytical threshold rooted from the variance of cross validation results.

The experimental results suggest that our method can be robustly used across many HPO optimizers. Depending on the user-defined thresholds, with 80\% to 90\% chance, our method achieves true regret within that threshold, saving unnecessary computation and reducing energy consumption. We also observe that overfitting exists in HPO even when cross-validation is used. We hope our work will draw the attention of the HPO community to the practical questions of how to set budget in advance and how to mitigate overfitting when tuning hyperparameters in machine learning.
\looseness-1

\section{Limitations and Broader Impact} 
%Automatic BO terminating is a rather under-explored topic. We empirically demonstrate the effectiveness of the proposed, however 
Further study of better variance estimate might be beneficial and we leave the modeling and estimation of this variance for the future work. In a broader scope of BO beyond HPO, a similar idea can be used when the point evaluations are not only corrupted by random noise but also adversarial corruptions (resulting again in a  discrepancy between the true objective and the computable target). 

In a broader context, we highlight that BO can reduce the computational cost required to tune ML models, mitigating the 
energy consumption and carbon footprint associated with brute force techniques such as random and grid search. The automatic termination criterion we presented in this work can have a positive societal impact by further reducing the cost of tuning ML models.

%\looseness-1 

\section{Acknowledgements} 
This research has been gratefully supported by NCCR Automation grant 51NF40 180545. The authors thank the anonymous reviewers of this paper for their helpful feedback. 

\bibliographystyle{apalike}
\bibliography{paper}

\begin{thebibliography}{}

\bibitem[Bayle et~al., 2020]{bayle2020crossvalidation}
Bayle, P., Bayle, A., Janson, L., and Mackey, L. (2020).
\newblock Cross-validation confidence intervals for test error.
\newblock In {\em Neural Information Processing Systems (NeurIPS)}.

\bibitem[Bergstra et~al., 2011]{bergstra2011algorithms}
Bergstra, J., Bardenet, R., Bengio, Y., and K{\'e}gl, B. (2011).
\newblock Algorithms for hyperparameter optimization.
\newblock In {\em Neural Information Processing Systems (NeurIPS)}.

\bibitem[Bergstra and Bengio, 2012]{JMLR:v13:bergstra12a}
Bergstra, J. and Bengio, Y. (2012).
\newblock Random search for hyperparameter optimization.
\newblock {\em Journal of Machine Learning Research}.

\bibitem[Chen et~al., 2018]{chen2018bayesian}
Chen, Y., Huang, A., Wang, Z., Antonoglou, I., Schrittwieser, J., Silver, D.,
  and de~Freitas, N. (2018).
\newblock {Bayesian optimization in AlphaGo}.

\bibitem[Dong and Yang, 2020]{dong2020nasbench201}
Dong, X. and Yang, Y. (2020).
\newblock Nas-bench-201: Extending the scope of reproducible neural
  architecture search.
\newblock In {\em International Conference on Learning Representations (ICLR)}.

\bibitem[Eriksson et~al., 2019]{NEURIPS2019_6c990b7a}
Eriksson, D., Pearce, M., Gardner, J., Turner, R.~D., and Poloczek, M. (2019).
\newblock Scalable global optimization via local {B}ayesian optimization.
\newblock In {\em Neural Information Processing Systems (NeurIPS)}.

\bibitem[Falkner et~al., 2018]{falkner2018bohb}
Falkner, S., Klein, A., and Hutter, F. (2018).
\newblock Bohb: Robust and efficient hyperparameter optimization at scale.
\newblock In {\em International Conference on Machine Learning}, pages
  1437--1446. PMLR.

\bibitem[Geisser, 1975]{geisser75}
Geisser, S. (1975).
\newblock The predictive sample reuse method with applications.
\newblock {\em Journal of The American Statistical Association}.

\bibitem[Ha et~al., 2019]{ha_unknown_search_bo}
Ha, H., Rana, S., Gupta, S., Nguyen, T., Tran-The, H., and Venkatesh, S.
  (2019).
\newblock {B}ayesian optimization with unknown search space.
\newblock In {\em Neural Information Processing Systems (NeurIPS)}, pages
  11795--11804. Curran Associates, Inc.

\bibitem[Hennig and Schuler, 2012]{Hennig2012}
Hennig, P. and Schuler, C.~J. (2012).
\newblock Entropy search for information-efficient global optimization.
\newblock {\em Journal of Machine Learning Research}.

\bibitem[Kirschner et~al., 2020]{kirschner2020}
Kirschner, J., Bogunovic, I., Jegelka, S., and Krause, A. (2020).
\newblock Distributionally robust {B}ayesian optimization.
\newblock In {\em Conference on Artificial Intelligence and Statistics
  (AISTATS)}.

\bibitem[Klein et~al., 2017]{Klein2017LearningCP}
Klein, A., Falkner, S., Springenberg, J.~T., and Hutter, F. (2017).
\newblock Learning curve prediction with {B}ayesian neural networks.
\newblock In {\em ICLR}.

\bibitem[Klein and Hutter, 2019]{Klein2019Tabular}
Klein, A. and Hutter, F. (2019).
\newblock Tabular benchmarks for joint architecture and hyperparameter
  optimization.
\newblock {\em arXiv:1905.04970}.

\bibitem[Klein et~al., 2020]{klein2020model}
Klein, A., Tiao, L.~C., Lienart, T., Archambeau, C., and Seeger, M. (2020).
\newblock Model-based asynchronous hyperparameter and neural architecture
  search.
\newblock {\em arXiv preprint arXiv:2003.10865}.

\bibitem[Kushner, 1963]{kushner1963new}
Kushner, H.~J. (1963).
\newblock A new method of locating the maximum point of an arbitrary multipeak
  curve in the presence of noise.
\newblock {\em Joint Automatic Control Conference}.

\bibitem[Li et~al., 2017]{li2017hyperband}
Li, L., Jamieson, K., DeSalvo, G., Rostamizadeh, A., and Talwalkar, A. (2017).
\newblock Hyperband: A novel bandit-based approach to hyperparameter
  optimization.
\newblock {\em The Journal of Machine Learning Research}, 18(1):6765--6816.

\bibitem[Li et~al., 2020]{li2020system}
Li, L., Jamieson, K., Rostamizadeh, A., Gonina, E., Ben-Tzur, J., Hardt, M.,
  Recht, B., and Talwalkar, A. (2020).
\newblock A system for massively parallel hyperparameter tuning.
\newblock {\em Proceedings of Machine Learning and Systems}, 2:230--246.

\bibitem[Lorenz et~al., 2016]{lorenz2016stopping}
Lorenz, R., Monti, R.~P., Violante, I.~R., Faisal, A.~A., Anagnostopoulos, C.,
  Leech, R., and Montana, G. (2016).
\newblock {Stopping criteria for boosting automatic experimental design using
  real-time fMRI with Bayesian optimization}.
\newblock {\em arXiv:1511.07827}.

\bibitem[Makarova et~al., 2021]{makarova2021riskaverse}
Makarova, A., Usmanova, I., Bogunovic, I., and Krause, A. (2021).
\newblock {Risk-averse Heteroscedastic Bayesian optimization}.
\newblock In {\em Conference on Neural Information Processing Systems
  (NeurIPS)}.

\bibitem[McLeod et~al., 2018]{mcleod18a}
McLeod, M., Roberts, S., and Osborne, M.~A. (2018).
\newblock Optimization, fast and slow: optimally switching between local and
  {B}ayesian optimization.
\newblock In {\em International Conference on Machine Learning (ICML)}.

\bibitem[Melis et~al., 2018]{melis2018on}
Melis, G., Dyer, C., and Blunsom, P. (2018).
\newblock On the state of the art of evaluation in neural language models.
\newblock In {\em International Conference on Learning Representations (ICLR)}.

\bibitem[Mo{\v{c}}kus, 1975]{movckus1975}
Mo{\v{c}}kus, J. (1975).
\newblock On {B}ayesian methods for seeking the extremum.
\newblock In {\em Optimization Techniques}.

\bibitem[Nadeau and Bengio, 2003]{nadeau2003inference}
Nadeau, C. and Bengio, Y. (2003).
\newblock Inference for the generalization error.
\newblock In {\em Neural Information Processing Systems (NeurIPS)}.

\bibitem[Nguyen et~al., 2017]{Nguyen2017regret}
Nguyen, V., Gupta, S., Rana, S., Li, C., and Venkatesh, S. (2017).
\newblock Regret for expected improvement over the best-observed value and
  stopping condition.
\newblock In {\em Asian Conference on Machine Learning}.

\bibitem[Rasmussen and Williams, 2006]{rasmussen2006}
Rasmussen, C.~E. and Williams, C. K.~I. (2006).
\newblock {\em Gaussian Processes for Machine Learning}.
\newblock The MIT Press.

\bibitem[Snoek et~al., 2012]{snoek12}
Snoek, J., Larochelle, H., and Adams, R.~P. (2012).
\newblock Practical {B}ayesian optimization of machine learning algorithms.
\newblock In {\em Conference on Neural Information Processing Systems
  (NeurIPS)}.

\bibitem[Srinivas et~al., 2010]{srinivas10}
Srinivas, N., Krause, A., Kakade, S., and Seeger, M. (2010).
\newblock Gaussian process optimization in the bandit setting: No regret and
  experimental design.
\newblock In {\em International Conference on Machine Learning (ICML)}.

\bibitem[Stone, 1974]{stone1974cross}
Stone, M. (1974).
\newblock Cross-validatory choice and assessment of statistical predictions.
\newblock {\em Journal of the royal statistical society. Series B
  (Methodological)}.

\bibitem[Swersky et~al., 2014]{Swersky2014FreezeThawBO}
Swersky, K., Snoek, J., and Adams, R. (2014).
\newblock Freeze-thaw {B}ayesian optimization.
\newblock {\em ArXiv}, abs/1406.3896.

\bibitem[Tiao et~al., 2021]{pmlr-v139-tiao21a}
Tiao, L.~C., Klein, A., Seeger, M.~W., Bonilla, E.~V., Archambeau, C., and
  Ramos, F. (2021).
\newblock {BORE}: {B}ayesian optimization by density-ratio estimation.
\newblock In {\em International Conference on Machine Learning (ICML)}.

\end{thebibliography}

\appendix
\clearpage
\section{Appendix} \label{sec:appendix}

\subsection{Experiments setting}
\label{sec:setting}

\subsubsection{BO setting}
\label{app:bo}

We used an internal BO implementation where expected improvement (EI) together with Mat{`e}rn-52 kernel in the GP are used. The hyperparameters of the GP includes output noise, a scalar mean value, bandwidths for every input dimension, 2 input warping parameters and a scalar covariance scale parameter. The closest open-source implementations are GPyOpt using input warped GP \footnote{\url{https://github.com/SheffieldML/GPyOpt}} or AutoGluon BayesOpt searcher \footnote{\url{https://github.com/awslabs/autogluon}}. We maximize type II likelihood to learn the GP hyperparameters in our experiments. We ran 200 iterations sequentially for benchmarks with cross validation, 500 iterations for NAS-HPO-Bench, 200 for NAS-Bench-201.

\subsubsection{Algorithm}
We present the pseudo-code for the termination criterion in \cref{app:algo}.
	\begin{algorithm}[h]
    \centering
    \caption{BO for HPO with cross-validation and automatic termination} 
        \label{alg:hpo-early-stopping}
        \begin{algorithmic}[1]
            \REQUIRE Model $\algo_{\gamma}$ parametrized by $\gamma \in \Gamma$ , data $\{\D_1,\dots,\D_k\}$ for $k$-fold cross-validation, \\ acquisition function $\alpha(\gamma)$
            \STATE Initialize $y_t^* = +\infty$ and $G_t =\{\}$
            % \STATE Initialize $\VarGset=\{\}$ and $\Dgamma_0 = \{\}$
        	\FOR{$t = 1, 2, \hdots$}
                \STATE Sample $\gamma_t \in \operatorname{arg\,max}_{\gamma \in \Gamma} \alpha(\gamma)$
                \FOR{$i = 1, 2, \hdots, k$}
                 \STATE Fit the model $\algo_{\gamma_t }(\cdot ; \D_{-i})$, where $\D_{-i}=\cup_{j\neq i}\D_i$ 
                 \STATE Evaluate the fitted model $y_t^i = \frac{1}{|\D_i|}\underset{\mathrm{x}_i, \mathrm y_i \in \D_i}{\sum} \ell(\mathrm{y}_i, \algo_{\gamma_t }(\mathrm{x}_i, \D_{-i}))$
                \ENDFOR
                \STATE Calculate the sample mean $y_t = \frac{1}{k} \underset{k}{\sum} y_t^i,$
                \IF{$y_t \leq y_t^*$} 
                \STATE Update $y_t^* = y_t$ and  the current best $\gamma_t^* = \gamma_t$ 
                \STATE Calculate the sample variance $\VarCV = \frac{1}{k} \sum_i (y_t - y_t^i)^2 $
                \STATE Calculate the variance estimate $\VarG (\gamma_t^*)\approx \left( \frac{1}{k} + \frac{|\D_{i}|}{|\D_{-i}|} \right) \VarCV$ from \cref{eq:variance_estimation} 
                \ENDIF
                \STATE Update $G_t = G_{t-1} \cup \gamma_t$ and $y_{1:t} = y_{1:t-1} \cup y_t$
                % \STATE Update $\VarGset = \VarGset \cup \{\VarGt\}$ and $ \VarGincumbent= \text{median}(\VarGset)$ 
                \STATE Update $\vec{\sigma_t}, \vec{\mu_t}$ with \cref{eq:posterior}
                \STATE Calculate upper bound $\bar{\simpleregret}_t := \underset{\gamma \in \Dt}{\min}\  \ucb_t(\gamma)  - \underset{\gamma \in \Gamma}{\min} \ \lcb_t(\gamma)$ for simple regret from \cref{eq:regret_bound}
                \IF{ the condition $\bar{\simpleregret}_t  \leq \sqrt{\VarG}(\gamma_t^*)$ holds} \STATE \textbf{terminate BO loop} 
                \ENDIF
            \ENDFOR
        	\STATE {\bfseries Output:} $\gamma_t^* $
        \end{algorithmic}
        \label{app:algo}
    \end{algorithm}

\subsubsection{Search spaces for cross-validation experiments}
\label{app:searchspace}

XGBoost (XGB) and RandomForest (RF) are based on scikit-learn implementations and their search spaces are listed in Table \ref{tab:parameter-range}.

\begin{table}[h]
\centering
\caption{Search spaces description for each algorithm.  \label{tab:parameter-range}}
%\begin{adjustbox}{width=0.7\columnwidth,center}
\begin{tabular}{lrrr}
\toprule
tasks &  hyperparameter & search space & scale\\
\midrule
\multirow{9}{*}{XGBoost}& n\_estimators & [$2$, $2^9$] &  log\\
 & learning\_rate & [$10^{-6}$, $1$] & log\\
 & gamma & [$10^{-6}$, $2^6$] & log\\
 & min\_child\_weight & [$10^{-6}, 2^5$] & log\\
 & max\_depth & [$2$, $2^5$] & log\\
 & subsample & [$0.5$, $1$] & linear\\
 & colsample\_bytree & [$0.3$, $1$] & linear\\
 & reg\_lambda & [$10^{-6}$, $2$] & log\\
 & reg\_alpha & [$10^{-6}$, $2$] & log\\
 \hline
 \multirow{3}{*}{RandomForest}  &  n\_estimators & [$1$, $2^8$] &  log\\
 & min\_samples\_split &  [$0.01$, $0.5$] & log \\
  & max\_depth & [$1$, $5$] & log\\
\bottomrule
\end{tabular}
%\end{adjustbox}
\end{table}

\subsubsection{Datasets in cross validation experiments}
\label{app:dataset}

We list the datasets that are used in our experiments, as well as their characteristics and sources in Table \ref{tab:meta}. For each dataset, we first randomly draw 20\% as test set and for the rest, we use 10-fold cross validations for regression datasets and 10-fold stratified cross validation for classification datasets. The actual data splits depend on the seed controlled in our experiments. For a given experiment, all the hyperparameters trainings use the same data splits for the whole tuning problem. For the experiments without cross-validation, we use 20\% dataset as validation set and the rest as training set. 

\begin{table}[h]
\centering
%\begin{adjustbox}{width=0.7\columnwidth,center}
\begin{tabular}{llrrrl}
\toprule
dataset &    problem\_type &  n\_rows &  n\_cols &  n\_classes & source \\
\midrule
openml14            &  classification &    1999 &      76 &       10 &       openml \\
openml20            &  classification &    1999 &     240 &       10 &       openml \\
tst-hate-crimes     &  classification &    2024 &      43 &       63 &     data.gov \\
openml-9910         &  classification &    3751 &    1776 &        2 &       openml \\
farmads             &  classification &    4142 &       4 &        2 &          uci \\
openml-3892         &  classification &    4229 &    1617 &        2 &       openml \\
sylvine             &  classification &    5124 &      21 &        2 &       openml \\
op100-9952          &  classification &    5404 &       5 &        2 &       openml \\
openml28            &  classification &    5619 &      64 &       10 &       openml \\
philippine          &  classification &    5832 &     309 &        2 &     data.gov \\
fabert              &  classification &    8237 &     801 &        2 &       openml \\
openml32            &  classification &   10991 &      16 &       10 &       openml \\
openml34538         &      regression &    1744 &      43 &        - &       openml \\
tst-census          &      regression &    2000 &      44 &        - &     data.gov \\
openml405           &      regression &    4449 &     202 &        - &       openml \\
tmdb-movie-metadata &      regression &    4809 &      22 &        - &       kaggle \\
openml503           &      regression &    6573 &      14 &        - &       openml \\
openml558           &      regression &    8191 &      32 &        - &       openml \\
openml308           &      regression &    8191 &      32 &        - &       openml \\
\bottomrule
\end{tabular}
%\end{adjustbox}
\caption{\label{tab:meta} Datasets used in our experiments including their characteristics and sources.}
\end{table}

\subsection{Detailed results}
\label{app:tabres}

\begin{figure}[h!]
\centering
\subfloat[]{
   \includegraphics[width=0.45\textwidth]{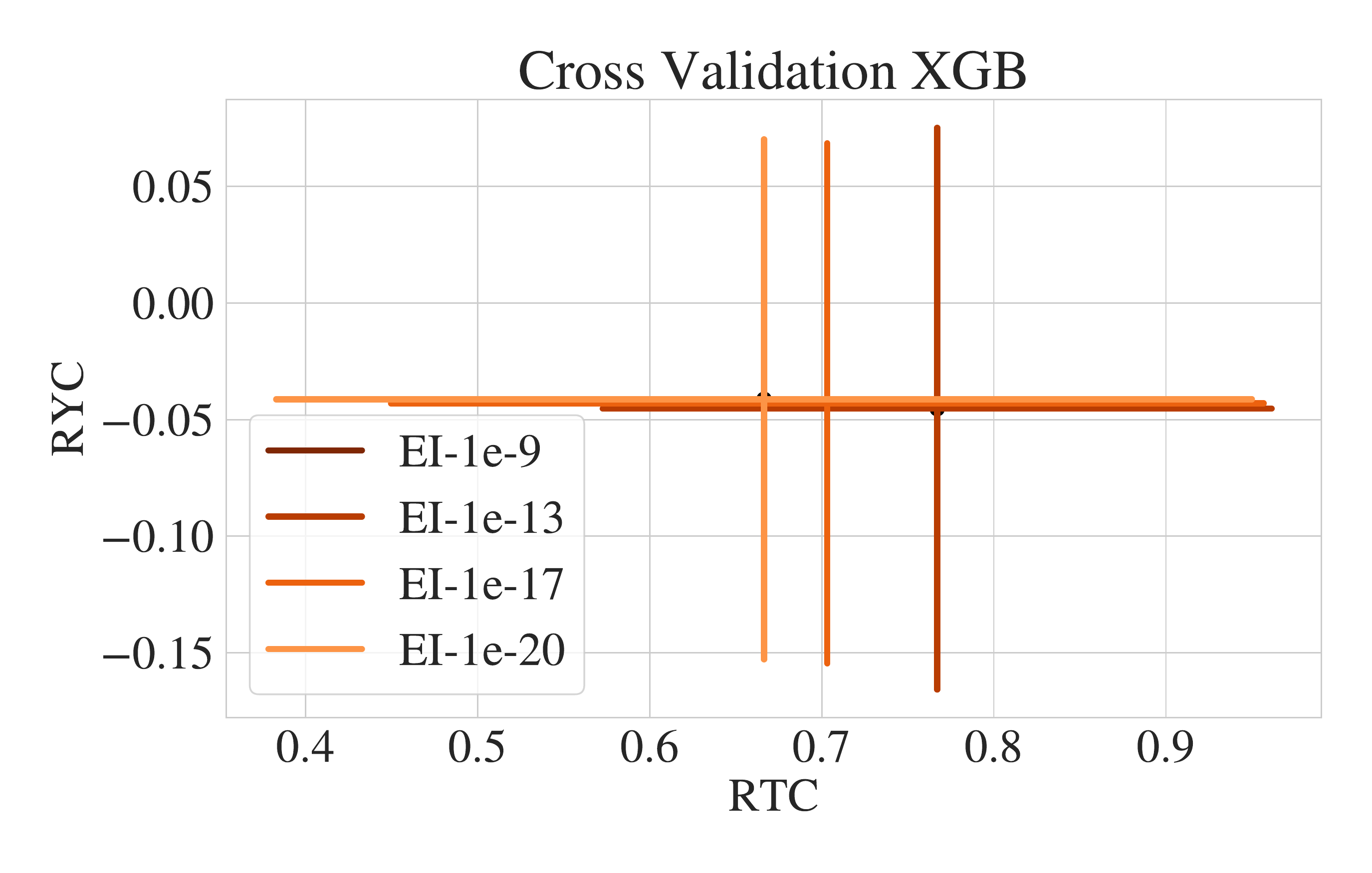}
  }  
    \hspace{1mm}
   \subfloat[]{
   \includegraphics[width=0.45\textwidth]{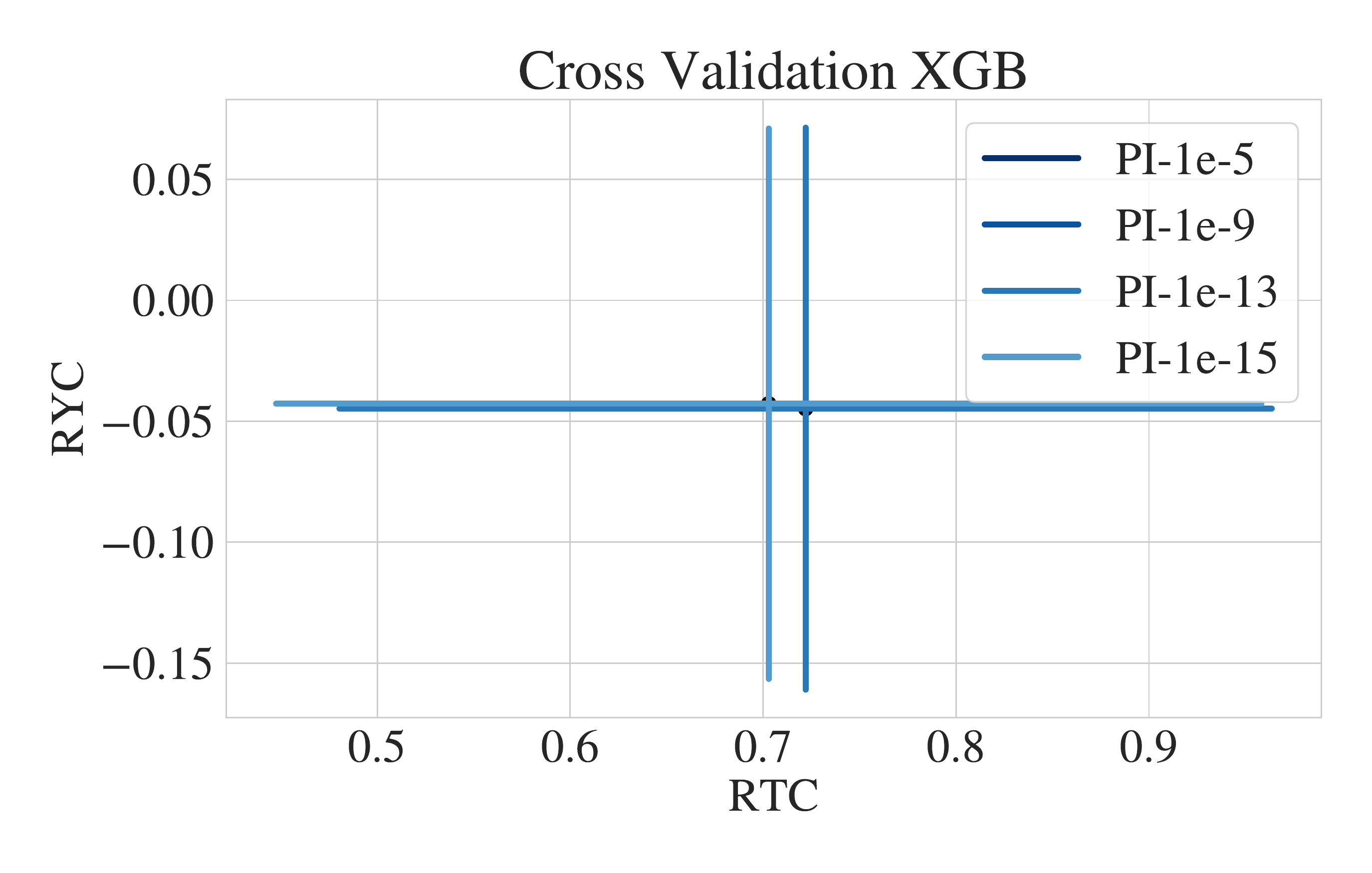}
  }
   % \hspace{1mm}  
  % \subfloat[]{
  %\label{fig:stop_regret}
  % \includegraphics[width=0.45\textwidth]{plots/true_regret_threshold.pdf}
  %}
  \caption{Detailed threshold study for the EI (left) and PI (right) baselines. Mean and standard deviation of RYC and RTC scores for EI and PI. \label{fig:bo-baselnes}}
\end{figure}

\begin{figure}[h!]
\centering
\subfloat[]{
   \includegraphics[width=0.45\textwidth]{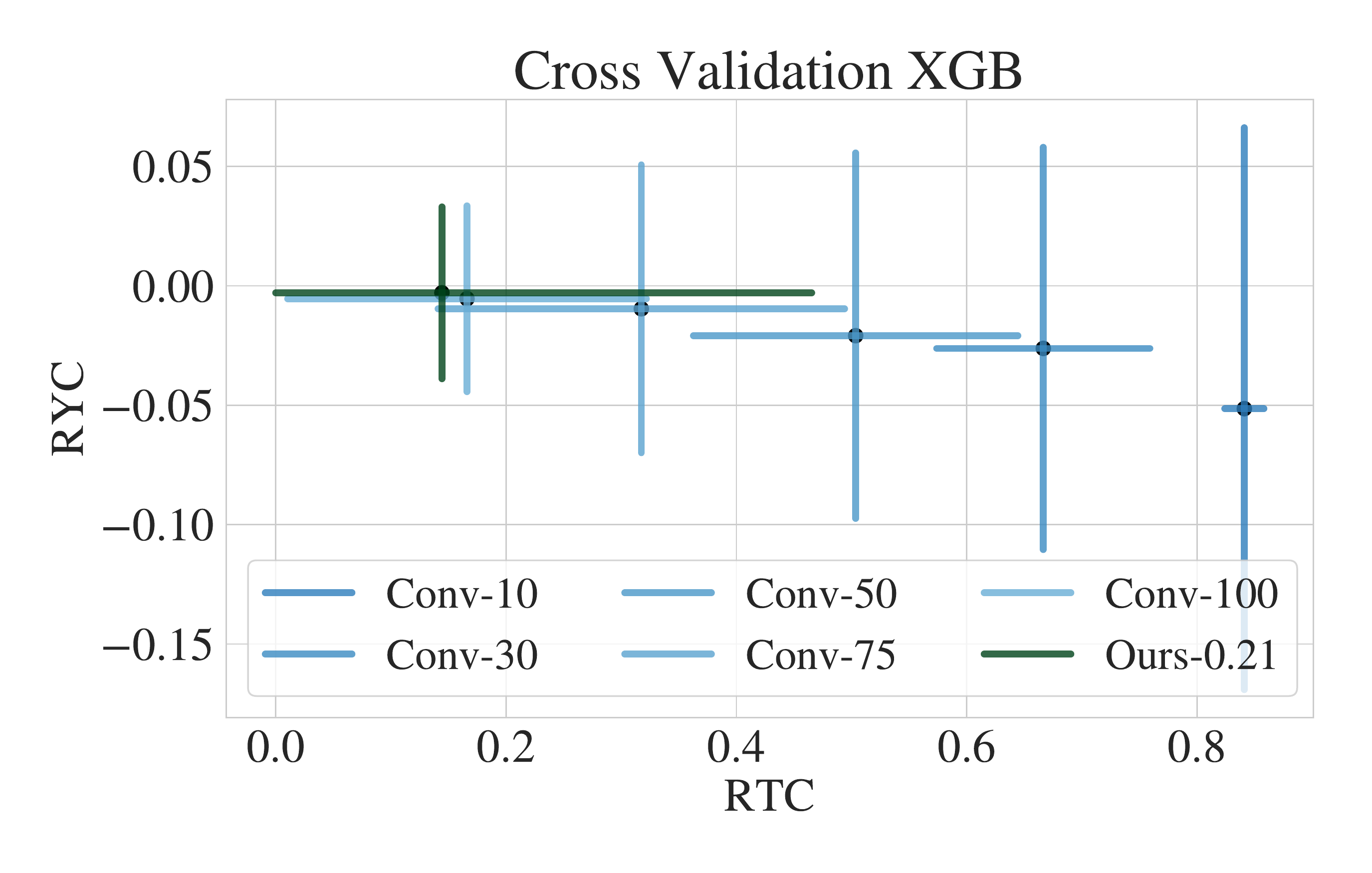}
  }  
    \hspace{1mm}
   \subfloat[]{
   \includegraphics[width=0.45\textwidth]{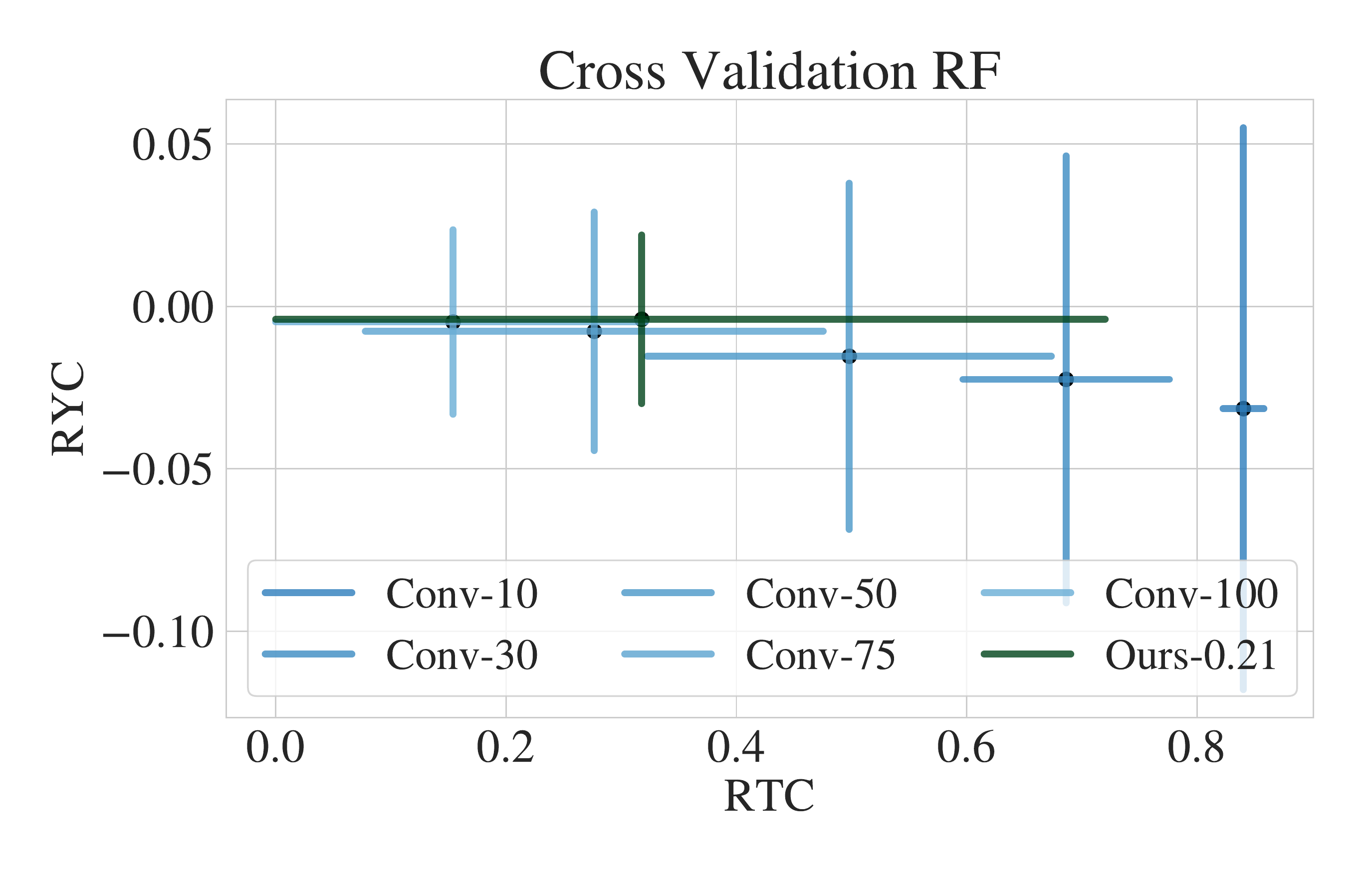}
  }
   \\
   \subfloat[]{
   \includegraphics[width=0.45\textwidth]{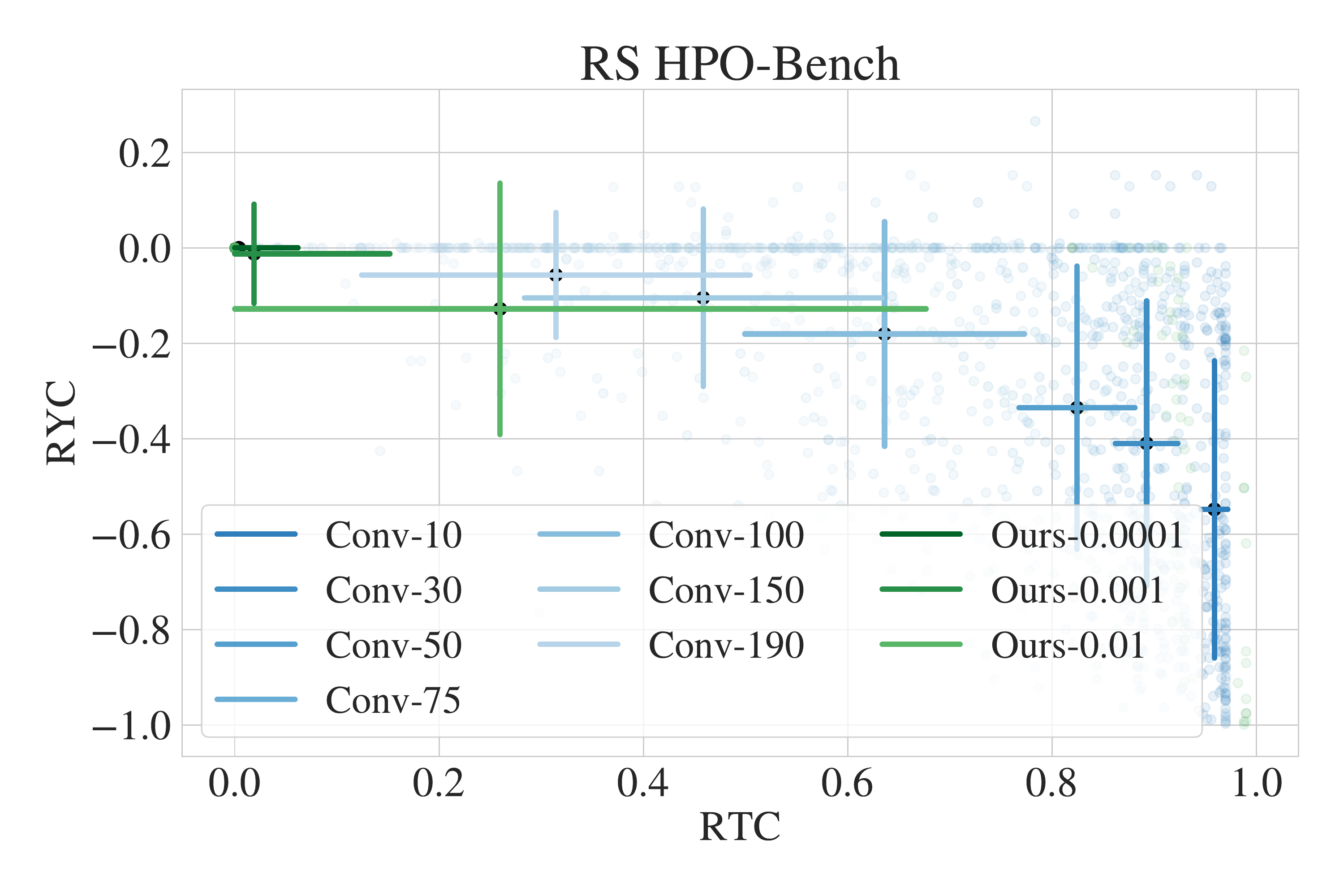}
  }  
    \hspace{1mm}
   \subfloat[]{
   \includegraphics[width=0.45\textwidth]{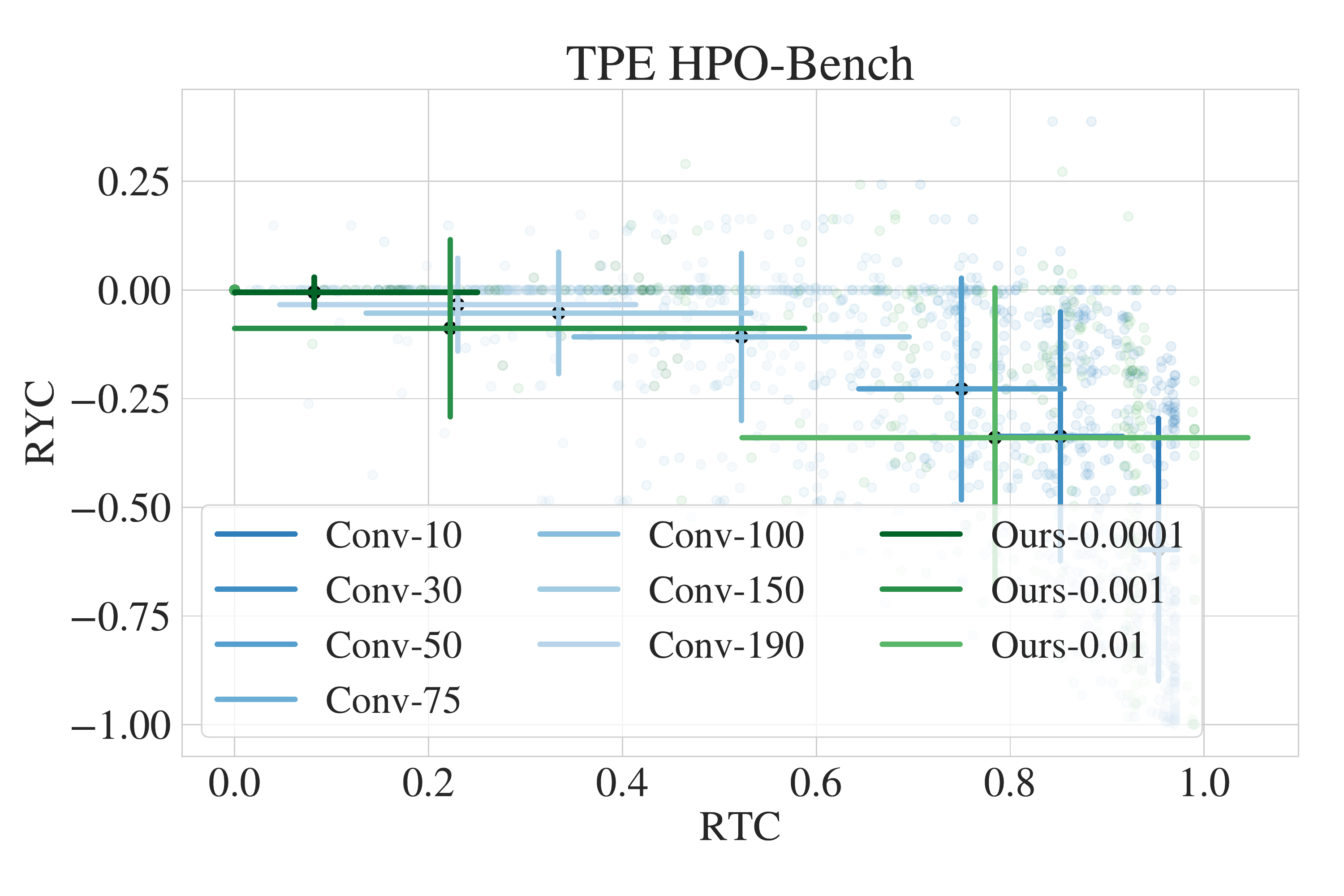}
  }
  
  \caption{Parameter $i$ study for Conv-$i$ baseline. The mean and standard deviation of RYC and RTC scores for (first line) HPO tuning XGB (a) and RF (b) using cross validation and (second line) for HPO-Bench datasets (c-d).  The Figure shows that even though there is an ideal for Conv-$i$, it changes not only across tasks and methods but even across different repetitions of a single method. The latter makes it particularly challenging to define a suitable $i$, since $i$ is a fixed, predetermined choice, that does not take the observed data into account (in contrast to our method that refines the regret estimation as the data is being observed). \label{fig:bo-baselnes-conv}}
\end{figure}

\cref{fig:bo-baselnes} demonstrates the results of threshold study EI and PI baselines for cross-validation. \cref{fig:bo-baselnes-conv} shows results for extended set of parameter $i$ for Conv-$i$ baseline. 

\begin{figure}[h!]
\centering
\subfloat[]{
   \includegraphics[width=0.45\textwidth]{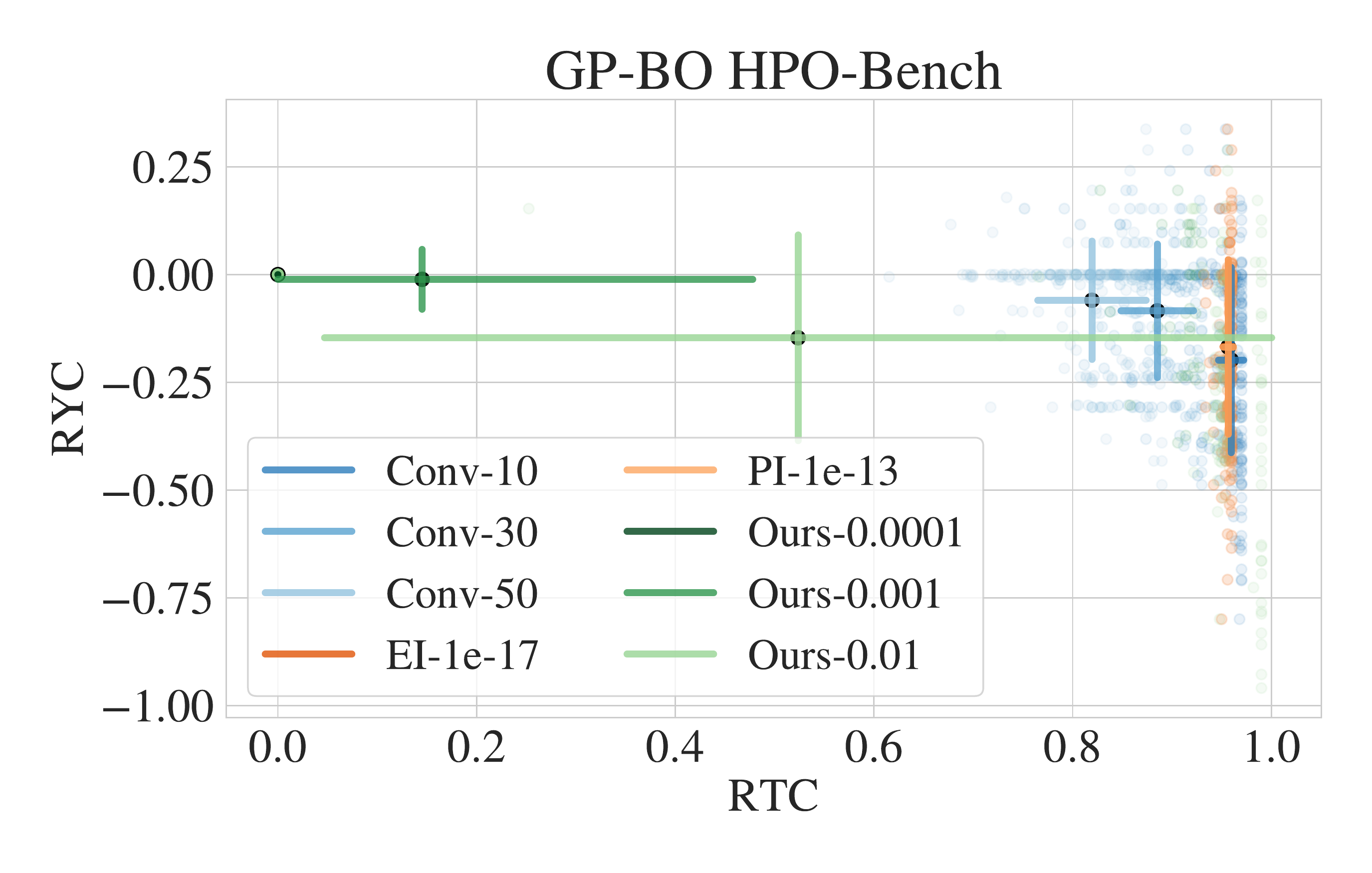}
  }  
    \hspace{1mm}
   \subfloat[]{
   \includegraphics[width=0.45\textwidth]{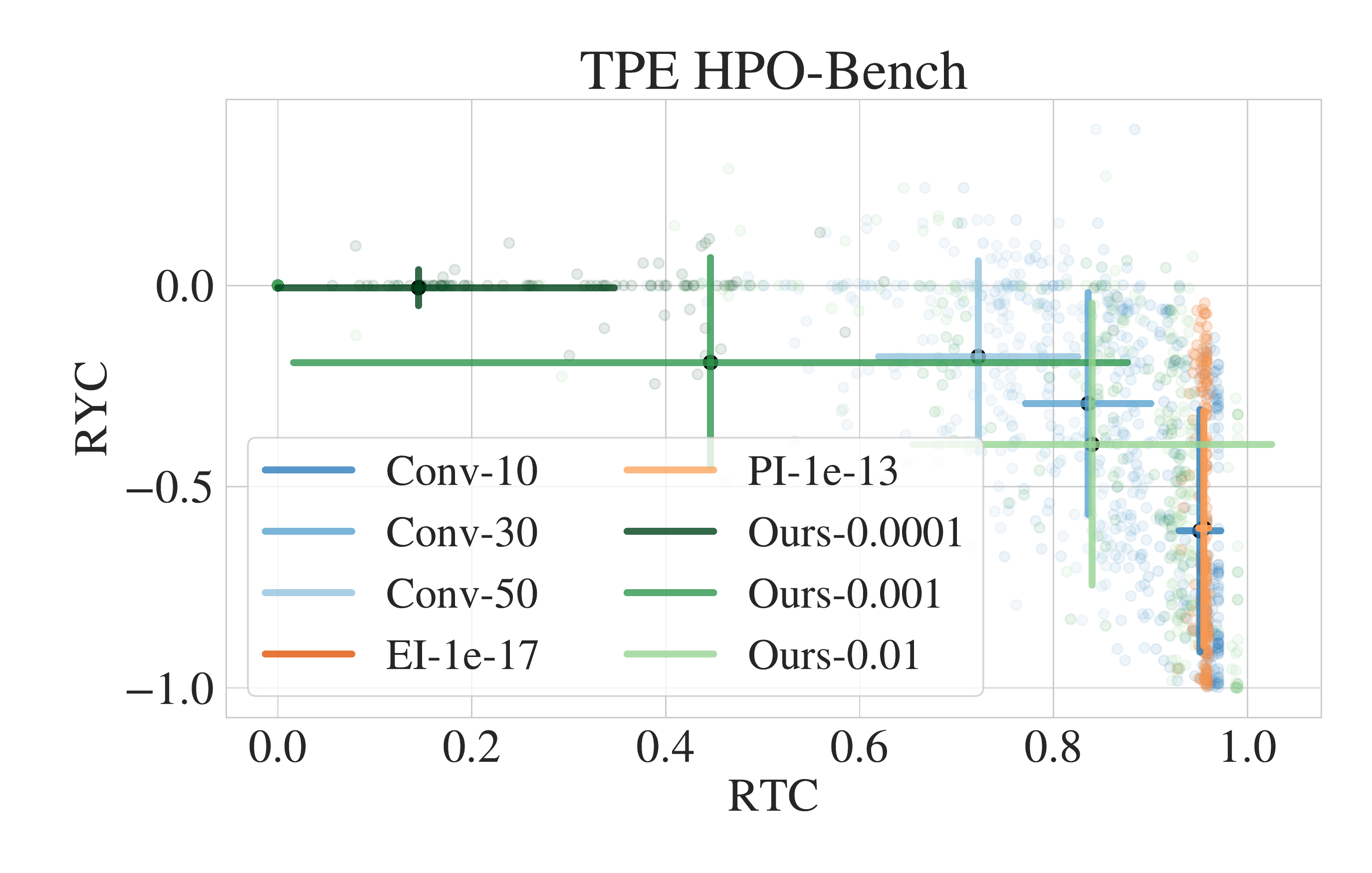}
  }
\\
 \subfloat[]{
   \includegraphics[width=0.45\textwidth]{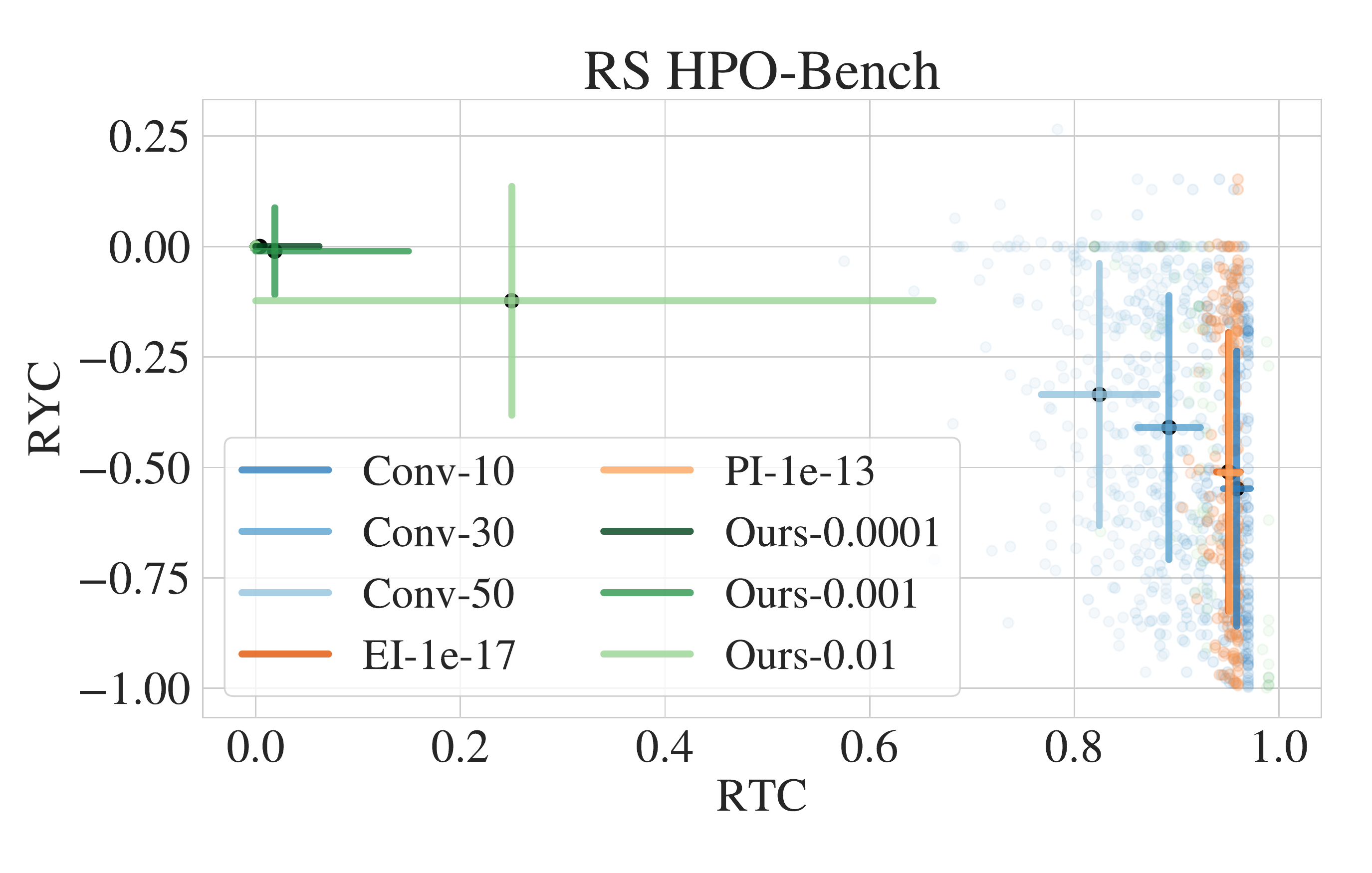}
  }    \hspace{1mm}
   \subfloat[]{
   \includegraphics[width=0.45\textwidth]{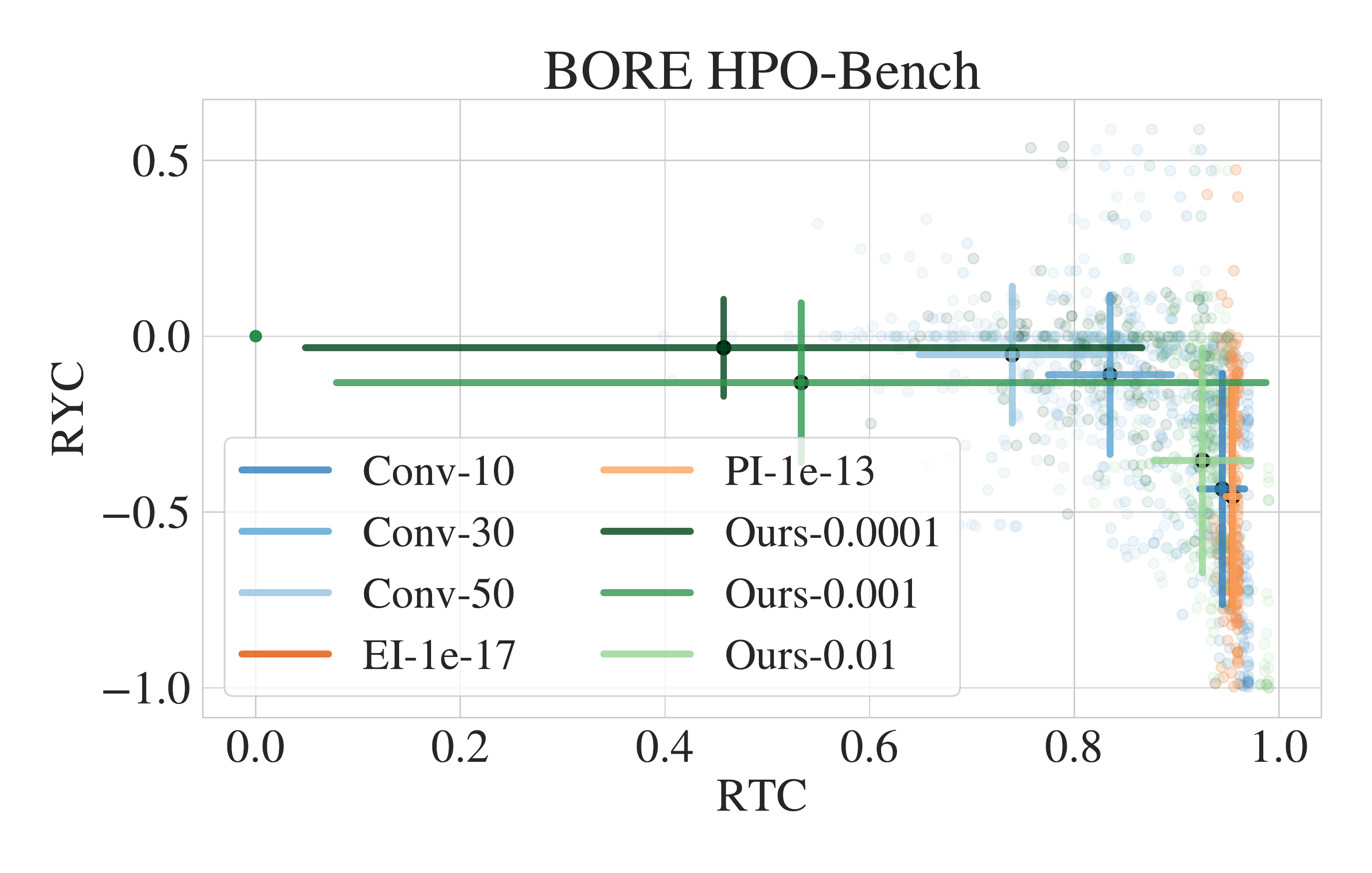}
  }
   % \hspace{1mm}  
  % \subfloat[]{
  %\label{fig:stop_regret}
  % \includegraphics[width=0.45\textwidth]{plots/true_regret_threshold.pdf}
  %}
  \caption{Fig. (a) - (d), the mean and standard deviation of RYC and RTC scores for considered automatic termination methods on HPO-Bench datasets using GP based BO (GP-BO), Random Search (RS), TPE and BORE optimizers. The mean value is shown in the big dot and the standard deviation is shown as error bar in both dimensions. \label{fig:hpo-bench-opt}}
\end{figure}

\begin{figure}[h!]
\centering
\subfloat[]{
   \includegraphics[width=0.45\textwidth]{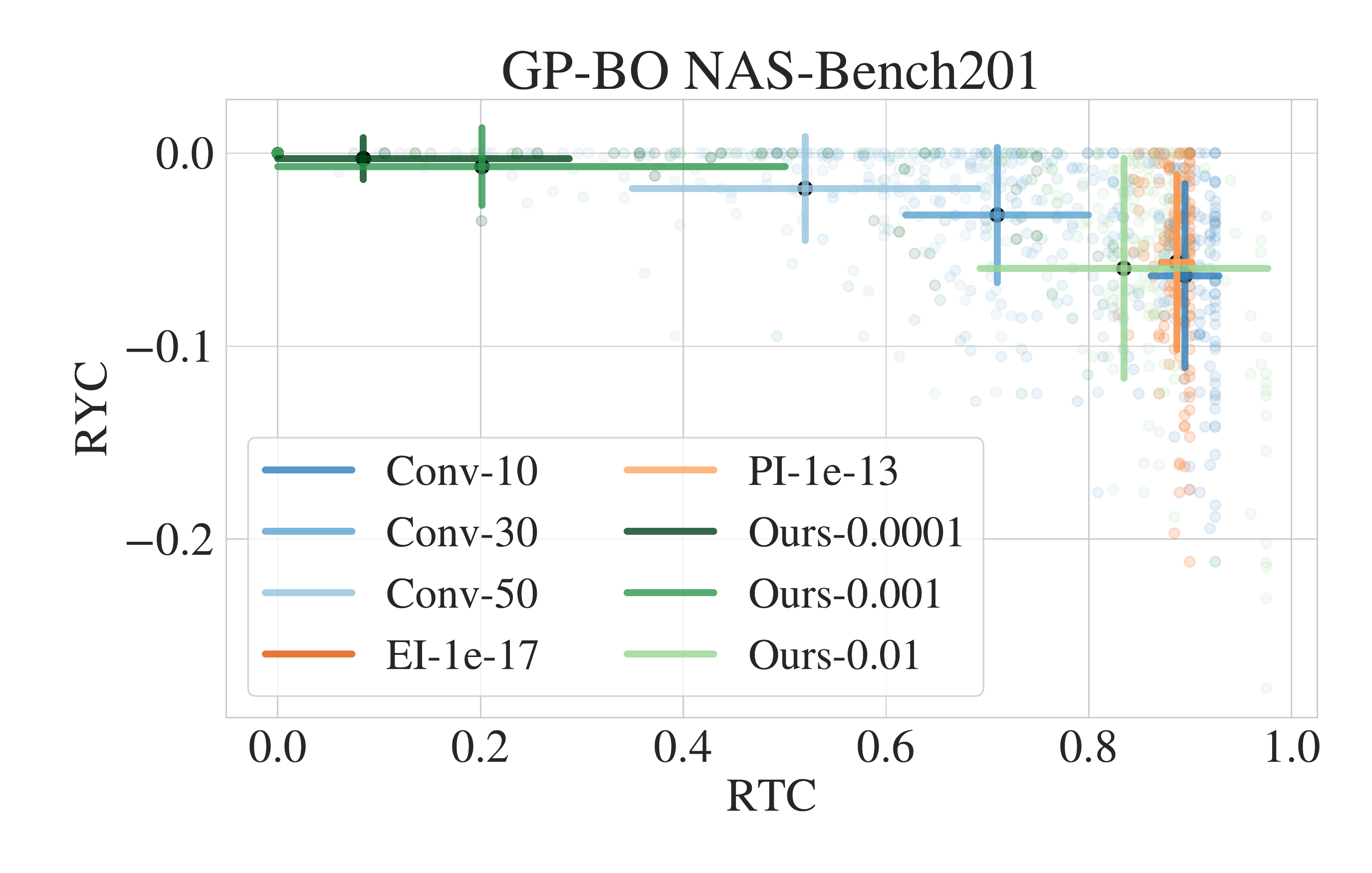}
  }  
    \hspace{1mm}
   \subfloat[]{
   \includegraphics[width=0.45\textwidth]{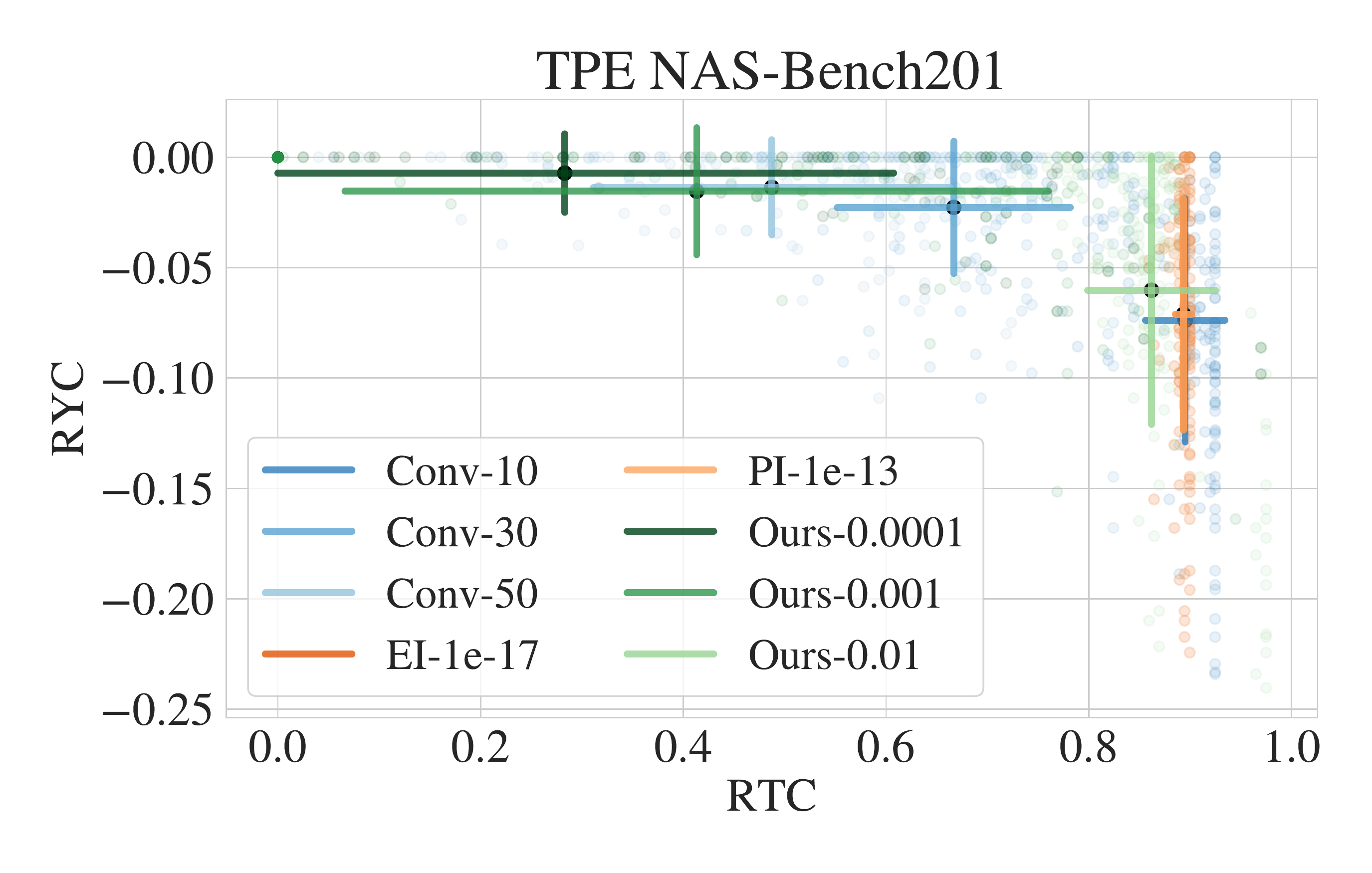}
  }
\\
 \subfloat[]{
   \includegraphics[width=0.45\textwidth]{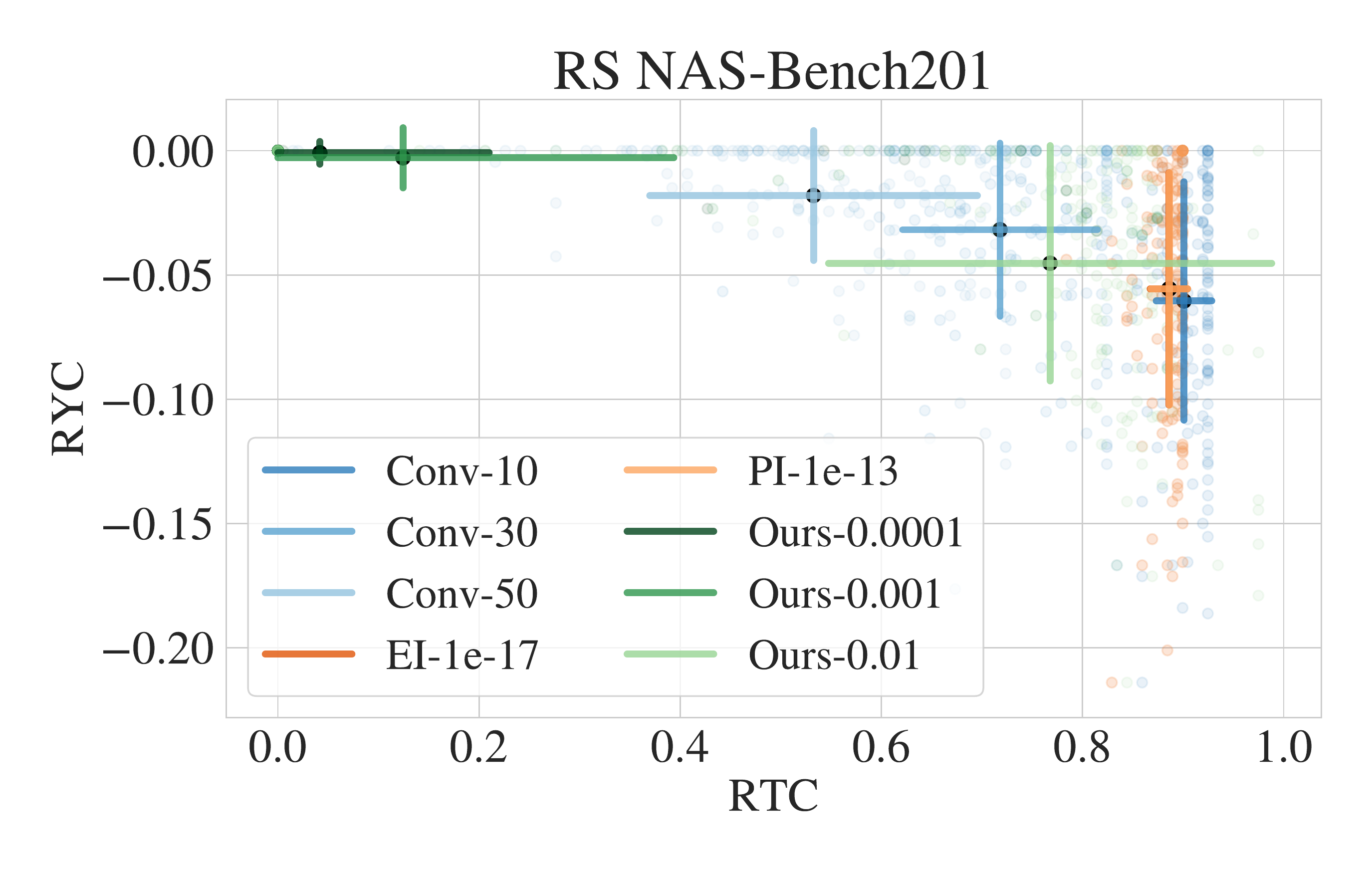}
  }    \hspace{1mm}
   \subfloat[]{
   \includegraphics[width=0.45\textwidth]{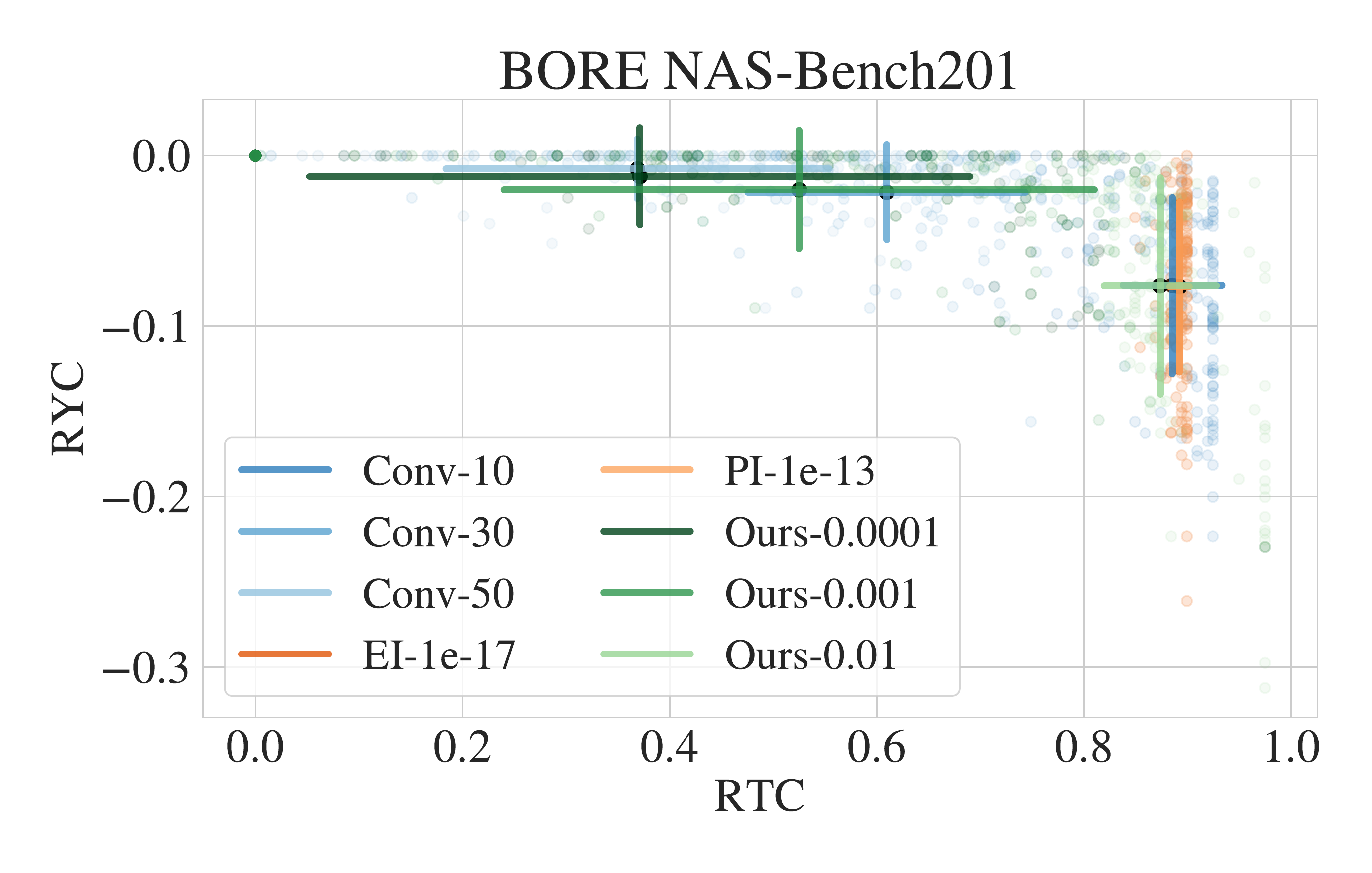}
  }
   % \hspace{1mm}  
  % \subfloat[]{
  %\label{fig:stop_regret}
  % \includegraphics[width=0.45\textwidth]{plots/true_regret_threshold.pdf}
  %}
  \caption{Fig. (a) - (d), the mean and standard deviation of RYC and RTC scores for considered automatic termination methods on NAS-Bench-201 datasets using GP based BO (GP-BO), Random Search (RS), TPE and BORE optimizers. The mean value is shown in the big dot and the standard deviation is shown as error bar in both dimensions. \label{fig:nas-bench-opt}}
\end{figure}

We also show the scatter plots of RTC and RYC scores for different automatic termination methods on HPO-Bench-datasets in \cref{fig:hpo-bench-opt} and the results on NAS-Bench-201 in \cref{fig:nas-bench-opt}.

\subsubsection{Detailed numbers of RYC and RTC scores}

We report detailed RYC scores and RTC scores of different HPO automatic termination methods for the experiments in the main text in \cref{tab:cv}, \cref{tab:hpo} and \cref{tab:nas201}.

\begin{table}[h]
\centering
%\begin{adjustbox}{width=0.7\columnwidth,center}
\begin{tabular}{lrrrr}
\toprule
{} & \multicolumn{2}{c}{RTC} & \multicolumn{2}{c}{RYC} \\
algo &     RF &    XGB &     RF &    XGB \\
\midrule
Conv\_10   &  0.840 &  0.841 & -0.031 & -0.051 \\
Conv\_30   &  0.686 &  0.666 & -0.022 & -0.026 \\
Conv\_50   &  0.498 &  0.504 & -0.015 & -0.021 \\
EI\_1e-08  &  0.896 &  0.850 & -0.057 & -0.052 \\
EI\_1e-12  &  0.895 &  0.779 & -0.055 & -0.047 \\
EI\_1e-16  &  0.893 &  0.718 & -0.052 & -0.045 \\
PI\_1e-4 &  0.898 &  0.875 & -0.059 & -0.059 \\
PI\_1e-08  &  0.895 &  0.814 & -0.055 & -0.052 \\
PI\_1e-12  &  0.894 &  0.739 & -0.055 & -0.044 \\
Ours\_0.21 &  0.318 &  0.144 & -0.004 & -0.003 \\
Ours\_0.5  &  0.580 &  0.224 & -0.013 & -0.006 \\
\bottomrule
\end{tabular}
%\end{adjustbox}
\caption{RTC and RYC scores for early stopping methods in cross validation benchmarks. \label{tab:cv}}
\end{table}

\begin{table}[h]
\centering
%\begin{adjustbox}{width=0.7\columnwidth,center}
\begin{tabular}{lrrrrrrrr}
\toprule
{} & \multicolumn{4}{c}{RTC} & \multicolumn{4}{c}{RYC} \\
dataset &  naval & parkinsons & protein &  slice &  naval & parkinsons & protein &  slice \\
\midrule
Conv\_10     &  0.943 &      0.947 &   0.946 &  0.942 & -0.605 &     -0.582 &  -0.117 & -0.432 \\
Conv\_30     &  0.826 &      0.837 &   0.837 &  0.840 & -0.064 &     -0.235 &  -0.021 & -0.119 \\
Conv\_50     &  0.748 &      0.729 &   0.734 &  0.747 & -0.038 &     -0.107 &  -0.008 & -0.058 \\
Ours\_0.0001 &  0.790 &      0.018 &   0.198 &  0.822 & -0.041 &     -0.012 &  -0.005 & -0.072 \\
Ours\_0.001  &  0.910 &      0.038 &   0.271 &  0.934 & -0.220 &     -0.031 &  -0.018 & -0.281 \\
Ours\_0.01   &  0.941 &      0.901 &   0.906 &  0.953 & -0.498 &     -0.378 &  -0.071 & -0.466 \\
\bottomrule
\end{tabular}
%\end{adjustbox}
\caption{RTC and RYC scores for early stopping methods in HPO-Bench. \label{tab:hpo}}
\end{table}

\begin{table}[h]
\centering
%\begin{adjustbox}{width=0.7\columnwidth,center}
\begin{tabular}{lrrrrrr}
\toprule
{} & \multicolumn{3}{c}{RTC} & \multicolumn{3}{c}{RYC} \\
dataset & ImageNet & cifar10 & cifar100 & ImageNet & cifar10 & cifar100 \\
\midrule
Conv\_10     &          0.880 &         0.889 &    0.888 &         -0.034 &        -0.098 &   -0.097 \\
Conv\_30     &          0.612 &         0.611 &    0.606 &         -0.010 &        -0.019 &   -0.036 \\
Conv\_50     &          0.372 &         0.361 &    0.372 &         -0.004 &        -0.006 &   -0.014 \\
Ours\_0.0001 &          0.274 &         0.311 &    0.519 &         -0.002 &        -0.008 &   -0.026 \\
Ours\_0.001  &          0.377 &         0.622 &    0.582 &         -0.005 &        -0.023 &   -0.033 \\
Ours\_0.01   &          0.837 &         0.902 &    0.879 &         -0.022 &        -0.106 &   -0.099 \\
\bottomrule
\end{tabular}
%\end{adjustbox}
\caption{RTC and RYC scores for early stopping methods in NAS-Bench-201. \label{tab:nas201}}
\end{table}

\subsubsection{Correlation between validation and test metrics}

\begin{figure}[h!]
\centering
 \subfloat[Training XGB on test-census dataset]{
 \label{fig:xgb-tst}
   \includegraphics[width=0.45\textwidth]{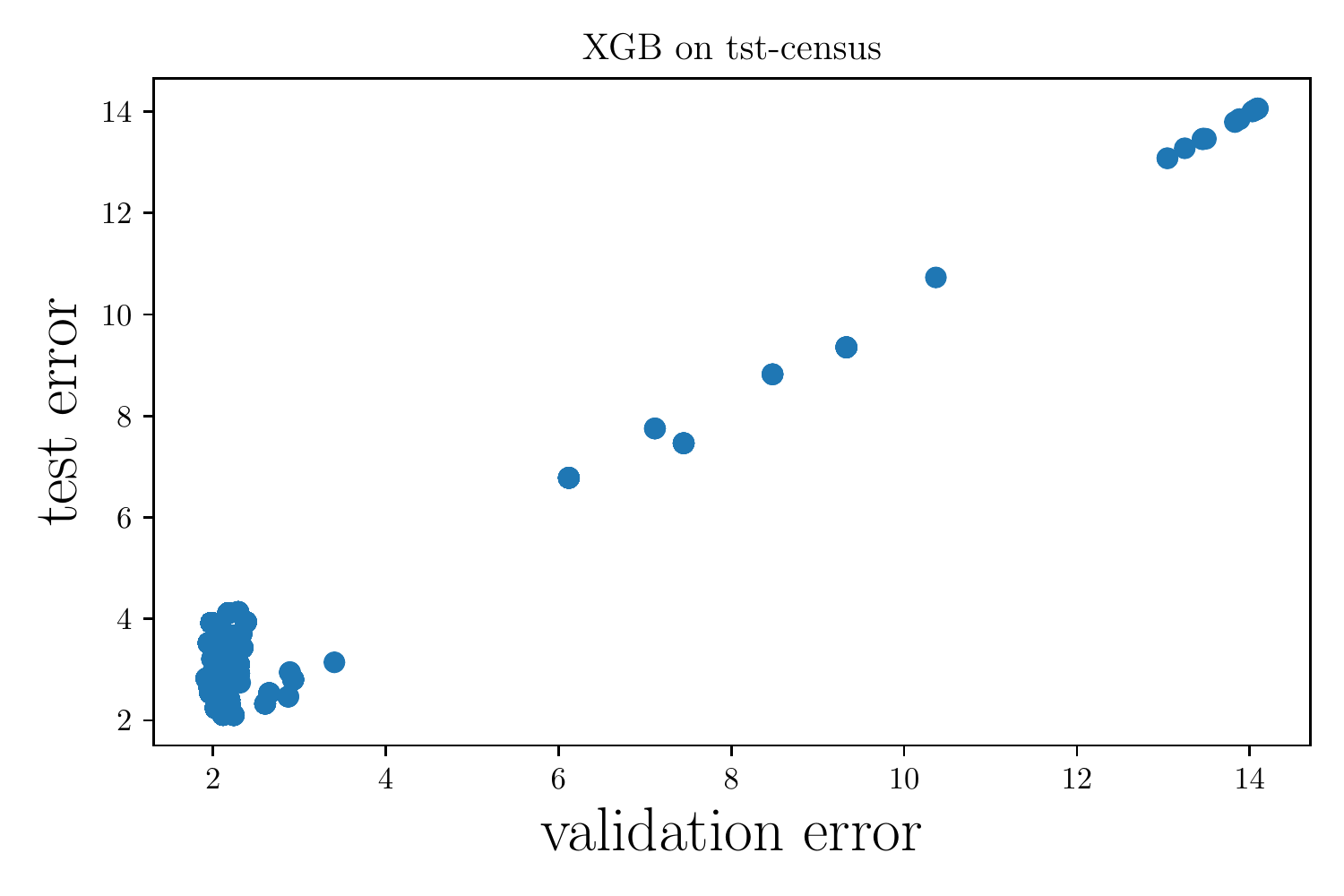}
 }
 \hspace{1mm}
 \subfloat[Training RF on test-census dataset.]{
 \label{fig:rf-tst}
   \includegraphics[width=0.45\textwidth]{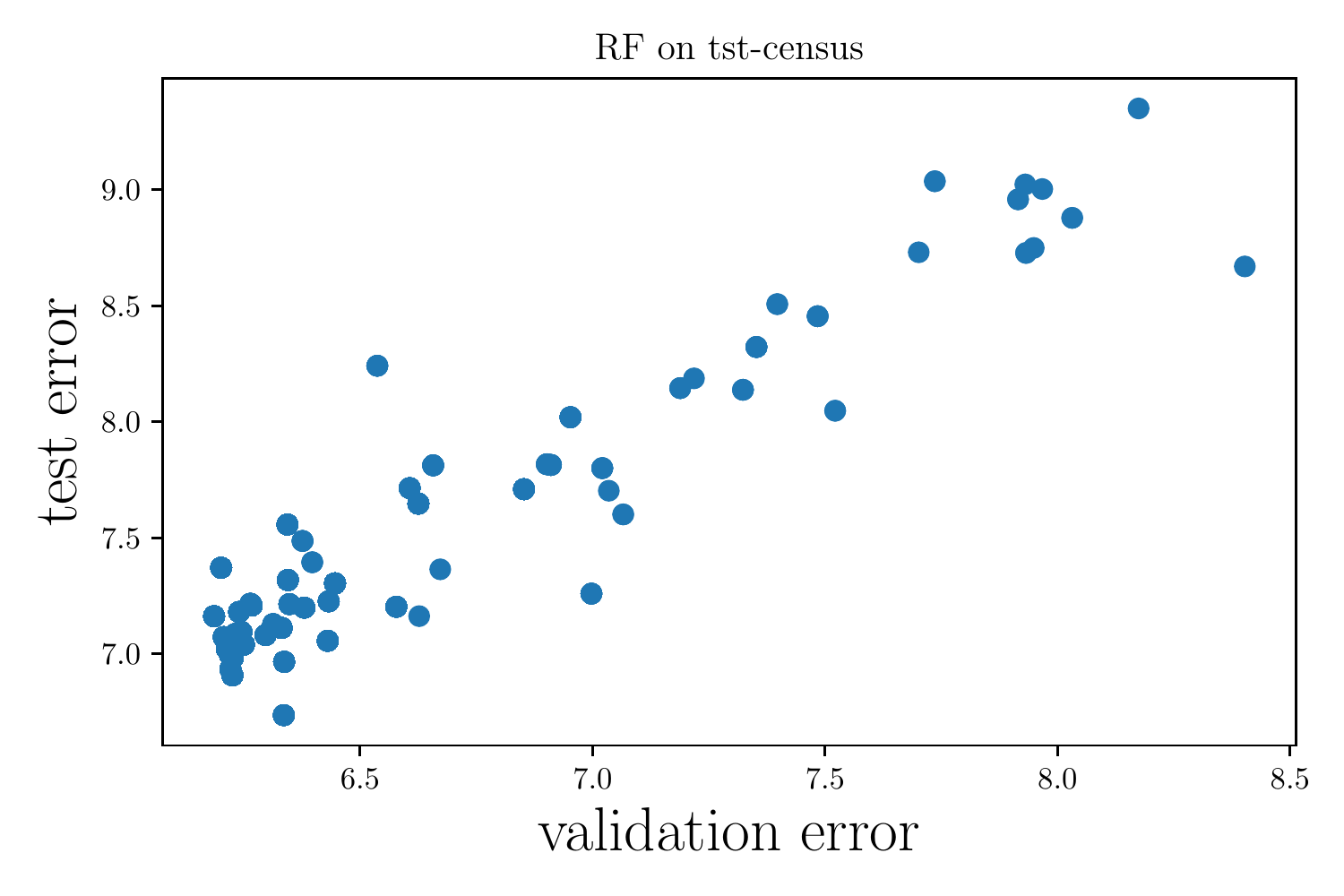}
  }
  \caption{We show validation error for training XGB (a) and RF (b) on \texttt{tst-census} dataset on the $x$-axis and test error on the $y$-axis. In the \textit{low} error region, the validation metrics are not well correlated with the test metrics. \label{fig:corr}}
\end{figure}

In \cref{fig:corr}, we show the correlation between validation and test metrics of hyperparameters when tuning XGB and RF on tst-census dataset in \cref{fig:corr}.

\subsubsection{The choice of $\beta_t$} \label{app:beta_choice}

High-probability concentration inequalities (aka confidence bounds) are important to reason about the unknown objective function and are used for theoretically grounded convergence guarantees in some (GP-UCB-based) BO methods \citep{srinivas10,ha_unknown_search_bo,kirschner2020,makarova2021riskaverse}.  There, $\beta_t$ stands for the parameter that balances between exploration vs. exploitation and ensures the validity of the confidence bounds. The choice of $\beta_t$ is then guided by the assumptions made on the unknown objective, for example, the objective being a sample from a GP or the objective having the bounded norm in RKHS (more agnostic case used in Section 3). 

\begin{figure}[h!]
\centering
 \subfloat{
   \includegraphics[width=0.7\textwidth]{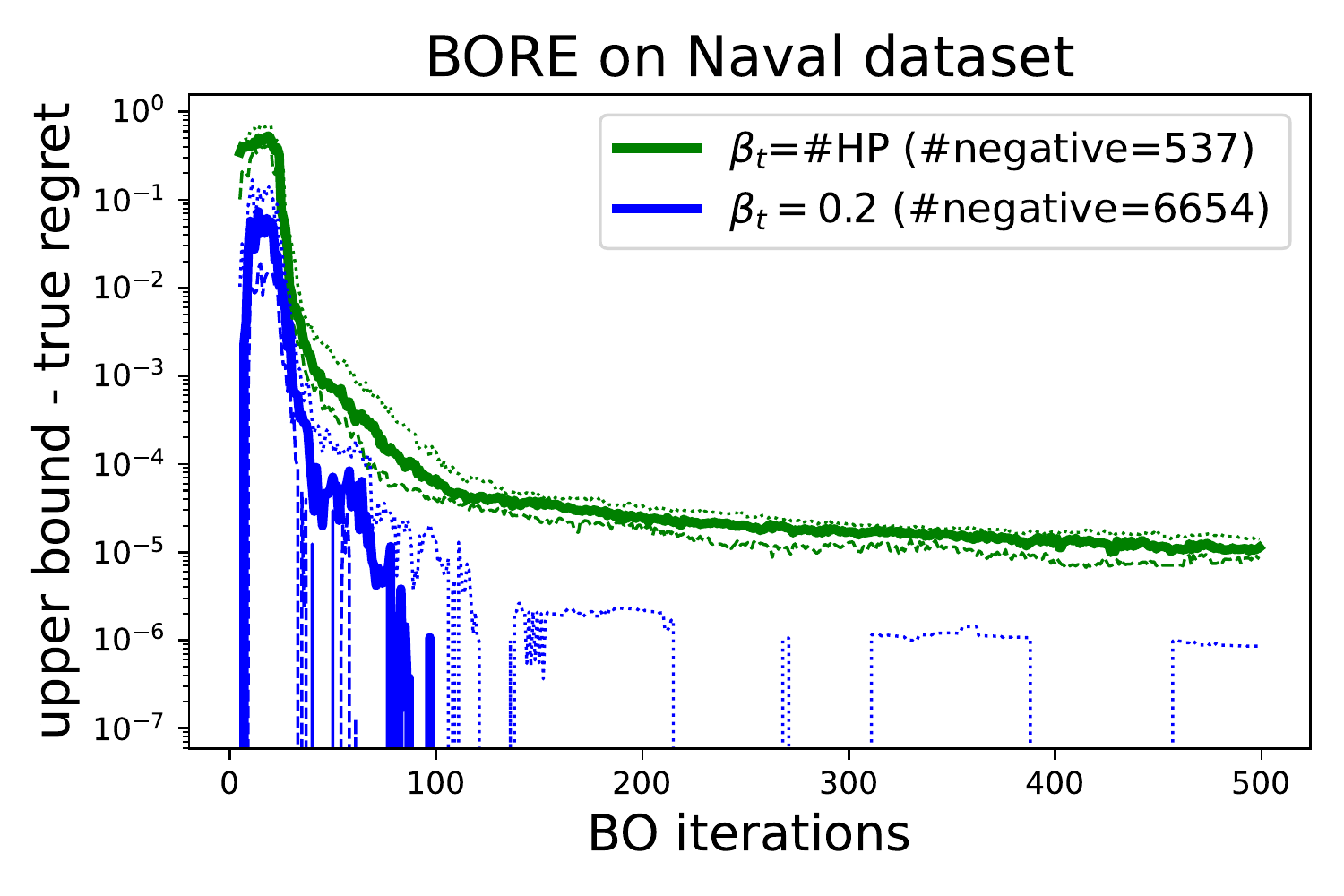}
 }
 \caption{The differences between upper bound and true regret for every BO iterations when using BORE to tune an MLP on the Naval dataset. The number of negative differences (the upper bound is smaller than the true regret) are shown in the legend next to the two options for computing $\beta_t$. \label{fig:betat}}
\end{figure}

In our experiments, we follow the common practice of scaling down $\beta_t$  which is usually used to improve performance over the (conservative) theoretically grounded values (see e.g., \cite{srinivas10,kirschner2020,makarova2021riskaverse}). Particularly, throughout this paper, we set $\beta_t = 2 \log (|\Gamma| t^2\pi^2/6\delta)$ where $\delta=0.1$ and $|\Gamma|$ is set to be the number of hyperparameters. We then further scale it down by a factor of 5 as defined in the experiments in \cite{srinivas10}. We provide an ablation study on the choice of $\beta_t$ in \cref{fig:betat}. 
%chosen to ensure the validity of confidence bounds. The choice of $\beta_t$ affects the upper bound estimation for the simple regret. We compare our choice of using the number of hyperparameters as the input size in Theorem 1 of \citet{srinivas10} to a common choice which is to use $\beta_t=0.2$. We use BORE optimize to tune an MLP on the Naval dataset and show the differences between upper bound and true regret for every BO iterations in \cref{fig:betat}. From the figure, it is clear that using $\beta_t=0.2$ underestimates the true regret much more frequently.

\clearpage
\section{Reproducibility Checklist}

% %
% For each question, change the default \verb|\answerTODO{}| (typeset \answerTODO)
% to
% \verb|\answerYes{[justification]}| (typeset \answerYes),
% \verb|\answerNo{[justification]}| (typeset \answerNo), or
% \verb|\answerNA{[justification]}| (typeset \answerNA).
% \textbf{You must include a brief justification to your answer,} either by
% referencing the appropriate section of your paper or providing a brief inline
% description. 
% For example:
% \begin{itemize}
% \item Did you include the license of the code and datasets? \answerYes{See
%   Section~\ref{sec:code}.}
% \item Did you include all the code for running experiments? \answerNo{We include
%   the code we wrote, but it depends on proprietary libraries for executing on a
%   compute cluster and as such will not be runnable without modifications. We
%   also include a runnable sequential version of the code that we also report
%   experiments in the paper with.}
% \item Did you include the license of the datasets? \answerNA{Our experiments
%   were conducted on publicly available datasets and we have not introduced new
%   datasets.}
% \end{itemize}
% Please note that if you answer a question with \verb|\answerNo{}|, we expect
% that you compensate for it (e.g., if you cannot provide the full evaluation
% code, you should at least provide code for a minimal reproduction of the main
% insights of your paper).

% Please do not modify the questions and only use the provided macros for your
% answers. Note that this section does not count towards the page limit. In your
% paper, please delete this instructions block and only keep the Checklist section
% heading above along with the questions/answers below.

\begin{enumerate}
\item For all authors\dots
  \begin{enumerate}
  \item Do the main claims made in the abstract and introduction accurately
    reflect the paper's contributions and scope?
    \answerYes{}
  \item Did you describe the limitations of your work?
    \answerYes{}
  \item Did you discuss any potential negative societal impacts of your work?
    \answerYes{We discuss any societal impacts in Section 6.}
  \item Have you read the ethics review guidelines and ensured that your paper
    conforms to them? \url{https://automl.cc/ethics-accessibility/}
    \answerYes{}
  \end{enumerate}
\item If you are including theoretical results\dots
  \begin{enumerate}
  \item Did you state the full set of assumptions of all theoretical results?
    \answerYes{}
  \item Did you include complete proofs of all theoretical results?
    \answerYes{}
  \end{enumerate}
\item If you ran experiments\dots
  \begin{enumerate}
  \item Did you include the code, data, and instructions needed to reproduce the
    main experimental results, including all requirements (e.g.,
    \texttt{requirements.txt} with explicit version), an instructive
    \texttt{README} with installation, and execution commands (either in the
    supplemental material or as a \textsc{url})?
    \answerYes{We provide the link to the code as footnote in Section 4.}
  \item Did you include the raw results of running the given instructions on the
    given code and data?
    \answerYes{}
  \item Did you include scripts and commands that can be used to generate the
    figures and tables in your paper based on the raw results of the code, data,
    and instructions given?
    \answerYes{}
  \item Did you ensure sufficient code quality such that your code can be safely
    executed and the code is properly documented?
    \answerYes{}
  \item Did you specify all the training details (e.g., data splits,
    pre-processing, search spaces, fixed hyperparameter settings, and how they
    were chosen)?
    \answerYes{}
  \item Did you ensure that you compared different methods (including your own)
    exactly on the same benchmarks, including the same datasets, search space,
    code for training and hyperparameters for that code?
    \answerYes{}
  \item Did you run ablation studies to assess the impact of different
    components of your approach?
    \answerYes{}
  \item Did you use the same evaluation protocol for the methods being compared?
    \answerYes{}
  \item Did you compare performance over time?
    \answerYes{}
  \item Did you perform multiple runs of your experiments and report random seeds?
    \answerYes{}
  \item Did you report error bars (e.g., with respect to the random seed after
    running experiments multiple times)?
    \answerYes{}
  \item Did you use tabular or surrogate benchmarks for in-depth evaluations?
    \answerYes{} 
  \item Did you include the total amount of compute and the type of resources
    used (e.g., type of \textsc{gpu}s, internal cluster, or cloud provider)?
    \answerNo{The lower bound of the regret estimation is the most compute intensive step and we have already included them in the open source repository.}
  \item Did you report how you tuned hyperparameters, and what time and
    resources this required (if they were not automatically tuned by your AutoML
    method, e.g. in a \textsc{nas} approach; and also hyperparameters of your
    own method)?
    \answerYes{}
  \end{enumerate}
\item If you are using existing assets (e.g., code, data, models) or
  curating/releasing new assets\dots
  \begin{enumerate}
  \item If your work uses existing assets, did you cite the creators?
    \answerYes{}
  \item Did you mention the license of the assets?
    \answerNo{}
  \item Did you include any new assets either in the supplemental material or as
    a \textsc{url}?
    \answerYes{}
  \item Did you discuss whether and how consent was obtained from people whose
    data you're using/curating?
    \answerNo{These are publicly available assets.}
  \item Did you discuss whether the data you are using/curating contains
    personally identifiable information or offensive content?
    \answerNA{}
  \end{enumerate}
\item If you used crowdsourcing or conducted research with human subjects\dots
  \begin{enumerate}
  \item Did you include the full text of instructions given to participants and
    screenshots, if applicable?
    \answerNA{}
  \item Did you describe any potential participant risks, with links to
    Institutional Review Board (\textsc{irb}) approvals, if applicable?
    \answerNA{}
  \item Did you include the estimated hourly wage paid to participants and the
    total amount spent on participant compensation?
    \answerNA{}
  \end{enumerate}
\end{enumerate}

\end{document}